\pdfoutput=1

\documentclass[11pt]{article}

\usepackage{acl}
\usepackage{times}
\usepackage{latexsym}
\usepackage{comment}
\usepackage{bbm}
\usepackage{rotating}
\usepackage[utf8]{inputenc}
\usepackage{times}
\usepackage{tikz,pgfplots}
\usepackage{array, makecell, multirow}
\usepackage{latexsym}
\usepackage{rotating,tabularx,siunitx}
\definecolor{cYellow}{RGB}{255,255,3}
\definecolor{cBlue}{RGB}{69,123,157}
\definecolor{cRed}{RGB}{231,56,71}
\definecolor{cRed_1}{RGB}{191,30,46}
\definecolor{cGray}{RGB}{168,218,219}
\definecolor{cBlue_2}{RGB}{5,48,97}
\definecolor{cBlue_1}{RGB}{115,186,214}
\definecolor{cBlue_3}{RGB}{13,76,109}
\definecolor{cBlue_4}{RGB}{64,121,160}
\definecolor{cOrange}{RGB}{250,134,0}
\definecolor{cBlue_6}{RGB}{13,76,109}
\definecolor{cBlue_7}{RGB}{16,106,130}
\definecolor{cBlue_8}{RGB}{19,136,160}
\definecolor{cBlue_9}{RGB}{115,184,214}
\usepackage{float}
\usetikzlibrary{plotmarks}
\usepackage{graphicx}
\usepackage{subfigure}
\usepackage{caption}
\usepackage{booktabs}
\usepackage{multirow}
\usepackage{hyperref}
\usepackage{multicol}
\usepackage{siunitx}
\usepackage{adjustbox}
\usepackage{tabularx}
\usepackage{enumitem}
\usepackage{pythonhighlight}
\newcolumntype{P}[1]{>{\centering\arraybackslash}p{#1}}
\usepackage{siunitx}
\usepackage{soul}
\usepackage{xcolor}
\usepackage{colortbl}
\sethlcolor{cYellow}
\usepackage{tcolorbox}
\usepackage{pifont}
\tcbuselibrary{skins, breakable, theorems}

\usepackage[T1]{fontenc}

\usepackage[utf8]{inputenc}

\usetikzlibrary{shapes.geometric}

\usepackage{microtype}
\usepackage{inconsolata}
\pgfplotsset{
/pgfplots/my legend/.style={
legend image code/.code={
    \node[star,star point ratio=2.25,rotate=36,minimum size=12pt,
          inner sep=0pt,draw=cBlue_4,solid,fill=cBlue_4,fill opacity=0.3] {};
   }
  }
}

\pgfplotsset{
/pgfplots/my medium/.style={
legend image code/.code={
    \node[fill opacity=0.7,text=cOrange] {\large $\odot$};
   }
  }
}
\pgfplotsset{
/pgfplots/my easy/.style={
legend image code/.code={
    \node[fill opacity=0.7,text=cBlue_3] {\large $\oplus$};
   }
  }
}

\pgfplotsset{
/pgfplots/my hard/.style={
legend image code/.code={
    \node[fill opacity=0.7,text=cRed_1] {\large $\otimes$};
   }
  }
}

\usepackage{pgfplots}
\pgfplotsset{compat=newest}
\pgfdeclareplotmark{mystar}{
    \node[star,star point ratio=2.25,minimum size=12pt,rotate=36,
          inner sep=0pt,draw=cBlue_4,solid,fill=cBlue_4,fill opacity=0.3] {};
}
\pgfdeclareplotmark{my_easy}{
    \node[fill opacity=0.3,text=cBlue_3] {$\oplus$};
}
\pgfdeclareplotmark{my_medium}{
    \node[fill opacity=0.3,text=cOrange] {$\odot$};
}
\pgfdeclareplotmark{my_hard}{
    \node[fill opacity=0.3,text=cRed_1] {$\otimes$};
}
\usepackage{fdsymbol}

\title{Text-Tuple-Table: Towards Information Integration in Text-to-Table Generation via Global Tuple Extraction}

\author{Zheye Deng$^{\clubsuit}$, Chunkit Chan$^{\clubsuit}$, Weiqi Wang$^{\clubsuit}$, Yuxi Sun$^{\varheartsuit}$, Wei Fan$^{\clubsuit}$, \\ \bf Tianshi Zheng$^{\clubsuit}$, Yauwai Yim$^{\clubsuit}$, Yangqiu Song$^{\clubsuit}$ \\
  $^{\clubsuit}$Department of Computer Science and Engineering, HKUST, Hong Kong SAR, China\\
  $^{\varheartsuit}$School of Computer Science, Fudan University, Shanghai, China \\
  \texttt{\{zdengah, yqsong\}@cse.ust.hk}\\
}

\begin{document}

\maketitle
\begin{abstract}
The task of condensing large chunks of textual information into concise and structured tables has gained attention recently due to the emergence of Large Language Models (LLMs) and their potential benefit for downstream tasks, such as text summarization and text mining. 
Previous approaches often generate tables that directly replicate information from the text, limiting their applicability in broader contexts, as text-to-table generation in real-life scenarios necessitates information extraction, reasoning, and integration.
However, there is a lack of both datasets and methodologies towards this task.
In this paper, we introduce \textsc{LiveSum}, a new benchmark dataset created for generating summary tables of competitions based on real-time commentary texts. 
We evaluate the performances of state-of-the-art LLMs on this task in both fine-tuning and zero-shot settings, and additionally propose a novel pipeline called \textsc{T}\textsuperscript{3}(\underline{\textbf{T}}ext-\underline{\textbf{T}}uple-\underline{\textbf{T}}able) to improve their performances. 
Extensive experimental results demonstrate that LLMs still struggle with this task even after fine-tuning, while our approach can offer substantial performance gains without explicit training. 
Further analyses demonstrate that our method exhibits strong generalization abilities, surpassing previous approaches on several other text-to-table datasets.
Our code and data can be found at \url{https://github.com/HKUST-KnowComp/LiveSum}.
\end{abstract}

\section{Introduction}

\begin{figure}[t!]
    \centering
    \includegraphics[width=0.44\textwidth]{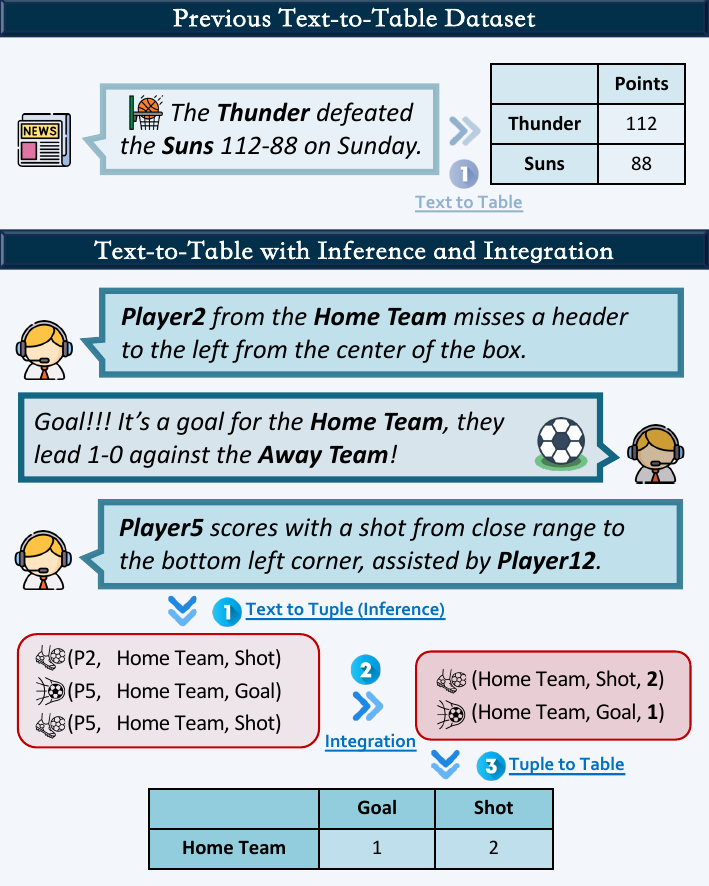}
    \vspace{-0.1in}
    \caption{An overview of the differences between our proposed \textsc{LiveSum} dataset and previous dataset~\cite{wiseman2017challenges}, as well as our proposed pipeline called \textsc{T}\textsuperscript{3}(\underline{\textbf{T}}ext-\underline{\textbf{T}}uple-\underline{\textbf{T}}able) which consists of three steps.}
    \vspace{-0.25in}
    \label{fig:intro}
\end{figure}

Reading extensive texts is demanding and time-consuming for humans, further compounded by the challenge of effectively capturing the key elements. 
Consequently, recent works have shifted to explore the structured summarization of text~\cite{jain2024structsum}, with tables being one highly prevalent form~\cite{wu2022text,li-etal-2023-sequence-sequence,sundar2024gtbls}.
These approaches improve text comprehension by extracting inherent yet valuable structural information from long unstructured text and enabling their applications in downstream scenarios, such as question answering~\cite{chen2020hybridqa,zhu2024tatllm}, text summarization~\cite{wiseman2017challenges,wang2020cord,Mulwad2023TowardsSE} and text data mining~\cite{li2023tablegpt,sui2024table}.

However, previous studies on text-to-table generation primarily rely on datasets traditionally used for table-to-text tasks~\cite{wiseman2017challenges,novikova-etal-2017-e2e}. 
One evident issue is that these tasks focus merely on format transformation, where the information in the table and the corresponding text representation are essentially similar~\cite{lebret-etal-2016-neural,bao2018table}. 
For example, in the upper part of Figure~\ref{fig:intro}, the table can be easily completed by extracting relevant numbers from the text without intermediate inference. 
Such seemingly meaningful correlations can introduce bias into the models, causing them to excel at replicating relevant information but struggle when it comes to categorizing and integrating numbers in complex scenarios.
This is further evidenced by the fact that fine-tuning models already perform very well and greatly surpass zero-shot LLMs~\cite{tang2023strucbench,sundar2024gtbls}. 
Hence, a more complicated dataset that requires information aggregation and minimizes the presence of spurious correlations, closely resembling real-world scenarios, is definitely needed for a more rigorous evaluation of models' text-to-table generation proficiency.

Apart from the research gap at the benchmark level, in terms of methodology, considerable attention has been given to studying the ability of LLMs to comprehend and generate complex structured outputs~\cite{tang2023strucbench,jain2024structsum}, driven by the exceptional success of LLMs in recent years~\cite{touvron2023llama,claude3,openai2024gpt4,DBLP:journals/corr/abs-2305-12870}. 
Extensive benchmarks indicate that LLMs exhibit sub-optimal performance in zero-shot settings, with multiple cases of generating inaccurate contents deviate from the given text~\cite{tang2023strucbench}. 
To address this issue, more sophisticated prompting mechanisms have been proposed~\cite{wei2022chain,khot2022decomposed,dua-etal-2022-successive}. 
Among them, \citet{jain2024structsum} introduce a divide-and-generate prompting approach to generate more accurate and informative tables, demonstrating its effectiveness in improving model performance. 
However, this simplistic approach of dividing text into paragraphs and generating tables is unsuitable in more complicated situations because table-relevant information may not be contiguous in the original text and may span across various paragraphs. 
Therefore, developing a robust prompting method is also needed for generating complex tables that capture crucial information from scattered text or paragraphs.

To resolve the aforementioned research gaps, we introduce a novel benchmark, \textsc{LiveSum},
which consists of 3,771 text-based live commentaries from real-world football matches, intending to evaluate the models' ability to generate summary tables.
Unlike previous benchmarks, our benchmark necessitates the model to possess the ability to extract correct and meaningful information from complex textual data, specifically emphasizing information integration, reasoning, and conceptualization skills~\cite{DBLP:conf/acl/WangFXBSC23,DBLP:conf/emnlp/WangF0XLSB23,DBLP:conf/acl/0001FLS0XWBLJCS24,DBLP:journals/corr/abs-2406-10885,DBLP:journals/corr/abs-2406-02106,DBLP:journals/corr/abs-2407-20564}. 
This is because commentaries in close temporal proximity or with similar semantic meanings may describe the same event, while verbs with similar meanings may refer to the same types of events. 
For example, in Figure~\ref{fig:intro}, the second and third dialogue boxes both describe the same goal event, and the verbs ``\emph{goal}'' and ``\emph{score}'' refer to the same goal event.

Along with the benchmark, we also introduce a robust prompting-based method \textsc{T}\textsuperscript{3} to address our proposed task.
Specifically, our method draws inspiration from the inherent attributes of the table, where each cell, along with its corresponding row header and column header, creates an informative triple (namely {\small \texttt{(row header, column header, cell)}}), which degenerates into a binary tuple when lacking row or column headers.
These tuples serve as cues for humans to locate specific information in the text and complete the table accordingly. 
Consequently, our pipeline begins by extracting the relevant tuples from the text, followed by the integration of these tuples, and ultimately generating one or more summary tables.

We hope that the proposed dataset, method, and experimental results can provide valuable insights for tasks such as text-to-table, as well as any task involving the generation of complex structured outputs from text. 
In summary, in this paper, we make the following contributions:
\vspace{-5pt}
\begin{itemize}
    \item To the best of our knowledge, \textsc{LiveSum} is the first benchmark dataset designed to evaluate the information integration ability of models in text-to-table generation tasks.
\vspace{-5pt}
    \item We introduce a novel \textsc{T}\textsuperscript{3}(\underline{\textbf{T}}ext-\underline{\textbf{T}}uple-\underline{\textbf{T}}able) prompting pipeline that functions as a flexible framework, applicable to any text-to-table generation tasks.
\vspace{-5pt}
    \item We conduct extensive experiments to evaluate the performance of LLMs under different settings and demonstrate that our \textsc{T}\textsuperscript{3} pipeline can bring significant improvements while showcasing excellent generalization capabilities.
\end{itemize}

\section{Task Definition}

We first provide a formal definition of the text-to-table generation. The input $\mathcal{S}$ consists of a textual passage with $n$ tokens, denoted as $\mathbf{x}=x_1,\dots,x_{n}$, and optionally, an instruction text with $m$ tokens, denoted as $\mathbf{y}=y_1,\dots,y_{m}$, which provides guidance on the format or content of the generated tables. The output $\mathcal{T}$ is a set of $k(k\geq1)$ tables, $\mathbf{T}^{1},\dots,\mathbf{T}^{k}$. For the output tables, we present a more detailed definition that covers two aspects: structure-related and content-related.
\paragraph{Structure}
We assume there are no merged cells in the tables for simplicity. Each table $\mathbf{T}^{i}$ has a caption $\mathbf{c}^{i}=c^{i}_1,\dots,c^{i}_{d}$, where $d=|\mathbf{c}^{i}|$, and it consists of $n^{i}_{r}$ rows and $n^{i}_{c}$ columns, resulting in a total of $n^{i}_{r} \times n^{i}_{c}$ cells. The cell $\mathbf{T}^{i}_{p,q}$ in the $p$-th row and $q$-th column is composed of a sequence of tokens: $\mathbf{T}^{i}_{p,q,1},\dots,\mathbf{T}^{i}_{p,q,r}$, where $r=|\mathbf{T}^{i}_{p,q}|$. The table $\mathbf{T}^{i}$ must have either a row header for all rows or a column header for all columns and it is also possible for the table to have both.
\paragraph{Content}
We define that the information in the output tables should be derived from the input $\mathbf{x}$ or can be inferred from $\mathbf{x}$. For each cell $\mathbf{T}^{i}_{p,q}$ in table $\mathbf{T}^{i}$, when combined with its row header $\mathbf{T}^{i}_{p,1}$ (if any), its column header $\mathbf{T}^{i}_{1,q}$ (if any), and the table's caption $\mathbf{c}^{i}$, it should convey the equivalent information as expressed in the input $\mathbf{x}$ and complies with the instruction $\mathbf{y}$ (if any).

\section{\textsc{LiveSum} Dataset}
\label{sec:dataset}

\begin{figure}[!t]
    \centering
    \includegraphics[width=0.42\textwidth]{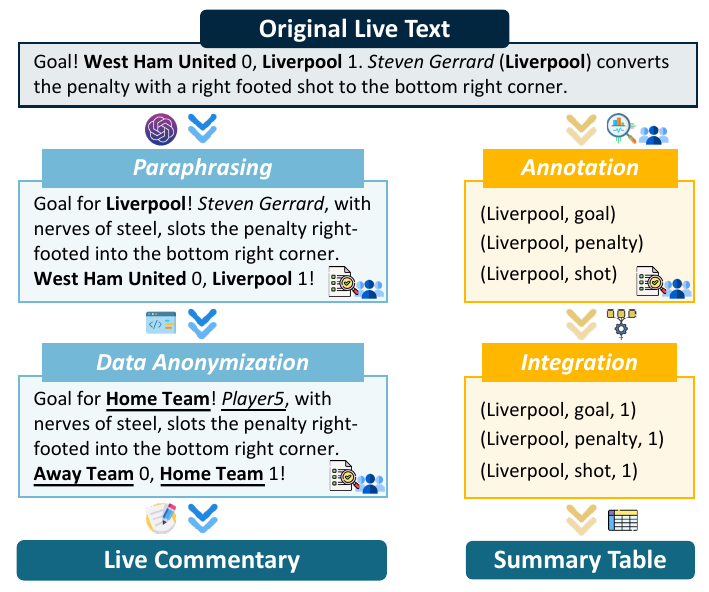}
    \vspace{-7pt}
    \caption{Overview of the pipeline for constructing the \textsc{LiveSum} dataset illustrated with a sample sentence.}
    \label{fig:construct_pipeline}
    \vspace{-14pt}
\end{figure}

We consider the problem of generating match statistic information tables from textual live commentary. Inspired by \textsc{RotoWire}~\cite{wiseman2017challenges}, a data-to-document dataset in the sports domain that aims to generate textual summaries by incorporating statistical data from basketball games, we instead focus on live commentary in football, which is available on BBC Sports\footnote{\url{https://www.bbc.com/sport/football}}. We crawl the data for the English Premier League from 2014 to 2023 and obtain complete commentary for 3,771 matches. 
Figure~\ref{fig:construct_pipeline} shows the pipeline we use to construct the dataset. Section~\ref{sec:construct_dataset_left} describes the generation process of the live commentary, and Section~\ref{sec:construct_dataset_right} describes the generation process for the summary table. More details are provided in Appendix~\ref{sec:app:construct_deil}.

\subsection{Live Commentary Generation}
\label{sec:construct_dataset_left}

To address formatting issues in the original textual live commentary on the website, we paraphrase the text to match the commentator's style while ensuring a certain degree of diversity.
Building upon previous studies~\cite{kim-etal-2023-soda, chen-etal-2023-places,DBLP:conf/coling/JiayangQC0SZ24,deng-etal-2023-gold,DBLP:conf/emnlp/ChengQCFWCRGZSZ23,DBLP:journals/corr/abs-2404-13627}, we employ ChatGPT~\cite{openai2022chatgpt} to generate complete live commentary automatically. Subsequently, to comply with privacy regulations and prevent bias in LLM benchmarks~\cite{DBLP:journals/corr/abs-2310-10383,DBLP:journals/corr/abs-2212-09292,DBLP:conf/acl/0003GLFH0CYYS24,DBLP:journals/corr/abs-2302-00539,DBLP:journals/corr/abs-2405-07667}, we also anonymize the data leveraging named entity recognition (NER) techniques~\cite{qi2020stanza} to produce the final version of the live commentary.

\begin{figure}[!t]
    \centering
    \includegraphics[width=0.42\textwidth]{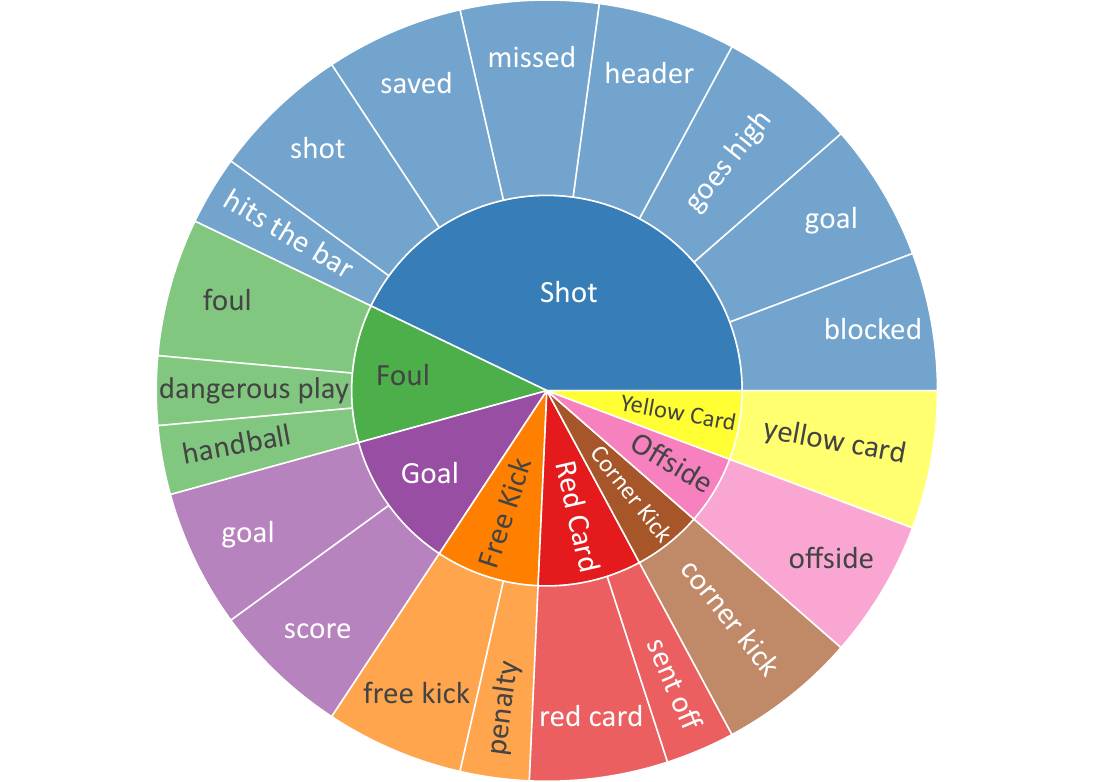}
    \caption{Eight types of event information (inner circle) that require summarization in \textsc{LiveSum} dataset, along with their common expressions (outer circle) in the commentary.}
    \vspace{-14pt}
    \label{fig:event_types}
\end{figure}

\subsection{Summary Table Generation}
\label{sec:construct_dataset_right}
 On the other hand, human annotators label a summary table for each match's commentary. We recruit five workers who are interested in football and are from English-speaking countries to perform the annotations. Since the occurrence of the events in football matches is deterministic, the ground truth is essentially unambiguous. In cases where there are inconsistencies in the annotated results, the correct answer is determined through a majority vote.

\subsection{Statistics}
\textsc{LiveSum} comprises a collection of 3,771 pairs, consisting of textual live commentaries and corresponding summary tables. We randomly split the entire dataset into training and test sets, resulting in 3,017 instances for the training set and 754 instances for the test set. On average, each live commentary segment consists of 1,256 words. \textsc{LiveSum} focuses on eight types of events, with the names and their corresponding common descriptions displayed in Figure~\ref{fig:event_types}.

\section{\textsc{T}\textsuperscript{3}(Text-Tuple-Table) Pipeline}
\label{sec:t3}

Our proposed \textsc{T}\textsuperscript{3}(\underline{\textbf{T}}ext-\underline{\textbf{T}}uple-\underline{\textbf{T}}able) pipeline is designed to mimic the intuitive steps followed by humans when performing this task.
When individuals aim to summarize a table from text, they typically extract pertinent or valuable tuples from the content, guided by any provided instructions, and then organize these tuples into one or more tables. Based on this concept, we divide this transformation into three stages: text-to-tuple, integration, and tuple-to-table, each of which is discussed in the following subsections. Taking Figure~\ref{fig:intro} as an example, we first extract key events mentioned in the text, then aggregate this information into consolidated tuples, and ultimately compile them into a table.

\vspace{-0.04in}
\subsection{Text-to-Tuple}
\begin{figure*}[!t]
\centering
    \begin{tikzpicture}[scale=0.72]

\begin{axis}[
    ybar,
    name=myGraph,
    ymin={0}, ymax={10},
    width={20cm}, height={7cm},
    bar width={12pt},
    ticks=both,
    ytick={0,1,2,3,4,5,6,7,8,9,10},
    ylabel={Root Mean Squared Error},
    ylabel style={yshift=-0.7em},
    ymajorgrids,
    xmajorgrids,
    enlarge x limits=0.048,
    clip = false,
    every tick label/.append style={font=\footnotesize},
    xtick = data,
    table/header=false,
    table/row sep=\\,
    xticklabels={},legend cell align={left},
    legend style={font=\small},
    legend style={at={(0.5, -0.48)}, anchor=north, legend columns=5},
]
\addplot[cBlue_1,fill=cBlue_1!40] table[x expr=\coordindex,y index=0]{4.564\\4.512\\4.287\\4.279\\4.162\\3.455\\3.613\\3.008\\2.911\\3.171\\2.918\\2.209\\2.139\\2.273\\2.225\\2.253\\2.079\\2.066\\1.631\\0.929\\0.438\\};
\draw[cRed_1, |<->|, line width=1.7pt] ([xshift = -0.03cm] axis description cs:0, 1.07) -- node[midway, rotate=0, text height=0.3cm, fill=cRed_1!10, yshift=12pt, text centered,  text depth=0.05cm, text=black] { {\textbf{Fine-Tune}} } ([xshift = 2.95cm] axis description cs:0, 1.07);
\draw[cRed_1, <->|,line width=2pt] ([xshift = 2.95cm] axis description cs:0, 1.07) -- node[midway, rotate=0, fill=cRed_1!10, yshift=12pt, text=black, text height=0.3cm, text centered,minimum height=2mm] { {\textbf{Zero-Shot}} } ([xshift = 14.7cm] axis description cs:0, 1.07);
\draw[cRed_1, <->|, line width=2pt] ([xshift = 14.7cm] axis description cs:0, 1.07) -- node[midway, rotate=0, text height=0.3cm, fill=cRed_1!10, yshift=12pt, text centered, text=black] {{\textbf{Zero-Shot w/ T\textsuperscript{3}}}} ([xshift = 18.43cm] axis description cs:0, 1.07);
\legend{\textbf{Root Mean Squared Error (Average)}\ };
\addlegendimage{my legend};
\addlegendentry{\textbf{Error Rate (Average)}\ };
\addlegendimage{my hard};
\addlegendentry{\textbf{Error Rate (Hard)}\ };
\addlegendimage{my medium};
\addlegendentry{\textbf{Error Rate (Medium)}\ };
\addlegendimage{my easy};
\addlegendentry{\textbf{Error Rate (Easy)}};
\end{axis}

\begin{axis}[
    axis y line*={right},
    axis x line*=center,
    ymin={0}, ymax={100},
    width={20cm}, height={7cm},
    ytick={0,10,20,30,40,50,60,70,80,90,100},
    ylabel style={yshift=0.5em},
    enlarge x limits=0.048,
    ylabel style={rotate=0},
    every tick label/.append style={font=\footnotesize},
    ylabel={Error Rate (\%)},
    xtick=data,
    xticklabel style={rotate=35,anchor=east, xshift=0.05cm, yshift=-0.3cm, font=\small},
    symbolic x coords={Mistral-7B-Instruct-v0.2 \textsc{(FT)},LLaMA-2-Chat 7B \textsc{(FT)},LLaMA-2-Chat 13B \textsc{(FT)},LLaMA-2-Chat 13B,LLaMA-2-Chat 13B \textsc{(CoT)},LLaMA-2-Chat 70B,LLaMA-2-Chat 70B \textsc{(CoT)},ChatGPT,ChatGPT \textsc{(CoT)},Claude 2.1,Claude 2.1 \textsc{(CoT)},Mistral Large,Mistral Large \textsc{(CoT)},GPT-4,GPT-4 \textsc{(CoT)},Claude 3 Opus,Claude 3 Opus \textsc{(CoT)},Claude 2.1 \textsc{(T\textsuperscript{3})},Mistral Large \textsc{(T\textsuperscript{3})},GPT-4 \textsc{(T\textsuperscript{3})},Claude 3 Opus \textsc{(T\textsuperscript{3})},},
]

\addplot[only marks,mark=mystar,
        point meta=y,
        ]  coordinates {
(Mistral-7B-Instruct-v0.2 \textsc{(FT)}, 76.030)
(LLaMA-2-Chat 7B \textsc{(FT)}, 75.910)
(LLaMA-2-Chat 13B \textsc{(FT)}, 75.71)
(LLaMA-2-Chat 13B, 75.33)
(LLaMA-2-Chat 13B \textsc{(CoT)}, 74.75)
(LLaMA-2-Chat 70B, 70.48)
(LLaMA-2-Chat 70B \textsc{(CoT)}, 71.40)
(ChatGPT, 61.03)
(ChatGPT \textsc{(CoT)}, 61.62)
(Claude 2.1, 57.16)
(Claude 2.1 \textsc{(CoT)}, 56.96)
(Mistral Large, 47.45)
(Mistral Large \textsc{(CoT)}, 47.12)
(GPT-4, 46.32)
(GPT-4 \textsc{(CoT)}, 46.20)
(Claude 3 Opus, 48.33)
(Claude 3 Opus \textsc{(CoT)}, 47.17)
(Claude 2.1 \textsc{(T\textsuperscript{3})}, 42.77)
(Mistral Large \textsc{(T\textsuperscript{3})}, 40.70)
(GPT-4 \textsc{(T\textsuperscript{3})}, 25.27)
(Claude 3 Opus \textsc{(T\textsuperscript{3})}, 14.04)
};

\addplot[only marks,mark=my_easy,mark size=3pt,cBlue_3,
        point meta=y,
        ]  coordinates {
(Mistral-7B-Instruct-v0.2 \textsc{(FT)}, 38.360)
(LLaMA-2-Chat 7B \textsc{(FT)}, 38.410)
(LLaMA-2-Chat 13B \textsc{(FT)}, 39.22)
(LLaMA-2-Chat 13B, 33.29)
(LLaMA-2-Chat 13B \textsc{(CoT)}, 31.83)
(LLaMA-2-Chat 70B, 12.34)
(LLaMA-2-Chat 70B \textsc{(CoT)}, 12.86)
(ChatGPT, 8.06)
(ChatGPT \textsc{(CoT)}, 10.61)
(Claude 2.1, 10.08)
(Claude 2.1 \textsc{(CoT)}, 14.06)
(Mistral Large, 0.27)
(Mistral Large \textsc{(CoT)}, 0.73)
(GPT-4, 4.64)
(GPT-4 \textsc{(CoT)}, 4.38)
(Claude 3 Opus, 2.52)
(Claude 3 Opus \textsc{(CoT)}, 1.59)
(Claude 2.1 \textsc{(T\textsuperscript{3})}, 8.95)
(Mistral Large \textsc{(T\textsuperscript{3})}, 8.82)
(GPT-4 \textsc{(T\textsuperscript{3})}, 3.18)
(Claude 3 Opus \textsc{(T\textsuperscript{3})}, 5.30)};

\addplot[only marks,mark=my_hard,mark size=3pt,cRed_1,
        point meta=y,
        ]  coordinates {
(Mistral-7B-Instruct-v0.2 \textsc{(FT)}, 95.520)
(LLaMA-2-Chat 7B \textsc{(FT)}, 95.400)
(LLaMA-2-Chat 13B \textsc{(FT)}, 94.42)
(LLaMA-2-Chat 13B, 93.29)
(LLaMA-2-Chat 13B \textsc{(CoT)}, 92.35)
(LLaMA-2-Chat 70B, 92.41)
(LLaMA-2-Chat 70B \textsc{(CoT)}, 94.24)
(ChatGPT, 90.62)
(ChatGPT \textsc{(CoT)}, 90.380)
(Claude 2.1, 90.58)
(Claude 2.1 \textsc{(CoT)}, 90.380)
(Mistral Large, 84.62)
(Mistral Large \textsc{(CoT)}, 84.08)
(GPT-4, 88.53)
(GPT-4 \textsc{(CoT)}, 88.73)
(Claude 3 Opus, 88.06)
(Claude 3 Opus \textsc{(CoT)}, 87.860)
(Claude 2.1 \textsc{(T\textsuperscript{3})}, 72.150)
(Mistral Large \textsc{(T\textsuperscript{3})}, 69.23)
(GPT-4 \textsc{(T\textsuperscript{3})}, 46.22)
(Claude 3 Opus \textsc{(T\textsuperscript{3})}, 21.29)
};

\addplot[only marks,mark=my_medium,mark size=3pt,cBlue_1!50,
        point meta=y,
        ]  coordinates {
(Mistral-7B-Instruct-v0.2 \textsc{(FT)}, 85.110)
(LLaMA-2-Chat 7B \textsc{(FT)}, 84.910)
(LLaMA-2-Chat 13B \textsc{(FT)}, 84.60)
(LLaMA-2-Chat 13B, 87.37)
(LLaMA-2-Chat 13B \textsc{(CoT)}, 87.42)
(LLaMA-2-Chat 70B, 88.59)
(LLaMA-2-Chat 70B \textsc{(CoT)}, 89.25)
(ChatGPT, 72.73)
(ChatGPT \textsc{(CoT)}, 72.75)
(Claude 2.1, 63.99)
(Claude 2.1 \textsc{(CoT)}, 61.700)
(Mistral Large, 52.450)
(Mistral Large \textsc{(CoT)}, 51.82)
(GPT-4, 46.05)
(GPT-4 \textsc{(CoT)}, 45.86)
(Claude 3 Opus, 51.36)
(Claude 3 Opus \textsc{(CoT)}, 49.60)
(Claude 2.1 \textsc{(T\textsuperscript{3})}, 44.99)
(Mistral Large \textsc{(T\textsuperscript{3})}, 42.37)
(GPT-4 \textsc{(T\textsuperscript{3})}, 25.830)
(Claude 3 Opus \textsc{(T\textsuperscript{3})}, 14.79)
};

\end{axis}

\end{tikzpicture}
\vspace{-0.05in}
\caption{The performance of various LLMs under fine-tune and zero-shot settings, as well as after the application of the \textsc{T}\textsuperscript{3} method on the test set of \textsc{LiveSum} dataset. The average RMSE and error rate for each model are displayed, along with the error rate for each of the three difficulty sections. More results are in Table~\ref{tab:main_result}.}
\vspace{-12pt}
\label{fig:main_result}
\end{figure*}

Considering the superior performance and flexibility of LLMs in information extraction compared to traditional techniques~\cite{ma-etal-2023-large,xu2023large}, we employ an LLM as our tuple extractor. We follow the instructions from InstructUIE~\cite{wang2023instructuie} and design the following prompting:
\vspace{-0.05in}
\begin{tcolorbox}[title={Text-to-Tuple Prompting Template}, colback = cBlue_1!10, colframe = cBlue_6,  coltitle=white,fonttitle=\bfseries\small, center title,fontupper=\small]
According to \texttt{<Instruction>}, please extract the relevant events and information in the form of tuples, structured as (\textit{subject}, \textit{object}, \textit{verb}) or (\textit{subject}, \textit{attribute}, \textit{value}): \texttt{<Text>}
\end{tcolorbox}
\vspace{-0.05in}
\noindent where \texttt{<Instrction>} is the directive for the current task, and \texttt{<Text>} is the text to be transformed.

\subsection{Information Integration}
\label{sec:t3_s2}
In this stage, we propose two approaches for integrating information. The first one involves direct execution by the LLM, using prompting to consolidate tuple data. The second one uses algorithms and code generated by the LLM to integrate tuple information, inspired by the LLMs' great success in code generation tasks~\cite{roziere2023code, luo2023wizardcoder, guo2024deepseekcoder}. \textsc{T}\textsuperscript{3} defaults to using code generation in this step. The promptings for these two methods are shown as follows:

\vspace{-0.05in}
\begin{tcolorbox}[title={Information Integration Prompting Template}, colback = cBlue_1!10, colframe = cBlue_7,  coltitle=white,fonttitle=\bfseries\small, center title,fontupper=\small,fontlower=\small]
\textbf{Direct Execution:}
According to \texttt{<Instruction>}, please integrate these tuples as required: \texttt{<Tuples>}
\tcblower
\textbf{Code Generation:}
According to \texttt{<Instruction>}, please develop an algorithm to consolidate these tuples as specified: \texttt{<Tuples>}
\end{tcolorbox}
\vspace{-0.05in}
\noindent where \texttt{<Instruction>} is the directive for the current task, and \texttt{<Tuples>} consists of the tuples extracted in the prior stage.

\subsection{Tuple-to-Table}

After obtaining the integrated tuples, we follow the previous implementation~\cite{tang2023strucbench,jain2024structsum} and use the following prompting to generate the final tables:

\vspace{-0.05in}
\begin{tcolorbox}[title={Tuple-to-Table Prompting Template}, colback = cBlue_1!10, colframe = cBlue_8,  coltitle=white,fonttitle=\bfseries\small, center title,fontupper=\small]
According to \texttt{<Instruction>}, please generate one or more tables based on the following tuples: \texttt{<Tuples>}
\end{tcolorbox}
\vspace{-0.05in}
\noindent where \texttt{<Instruction>} is the directive for the current task, and \texttt{<Tuples>} consists of the tuples produced in the prior stage.

\section{Experimental Setup}
\label{sec:experiment_setup}

\paragraph{Baseline Models}
In this study, we conduct fine-tuning on the \textsc{LiveSum} dataset using three representative open-source LLMs: Mistral-7B-Instruct-v0.2~\cite{jiang2023mistral}, LLaMA-2 Chat 7B and LLaMA-2 Chat 13B~\cite{touvron2023llama}. We fine-tune these models following the current state-of-the-art fine-tuning methodologies~\cite{tang2023strucbench}. Therefore, the outcomes represent the best results achievable with the present fine-tuning methods. We also evaluate eight state-of-the-art LLMs in zero-shot settings: LLaMA-2 Chat 13B, LLaMA-2 Chat 70B~\cite{touvron2023llama}, Mistral Large~\cite{mistral}, Claude 2.1~\cite{claude21}, Claude 3 Opus~\cite{claude3}, ChatGPT~\cite{openai2022chatgpt}, and GPT-4~\cite{openai2024gpt4}. Because of the limitation of the input lengths, we evaluate two LLMs in few-shot settings: ChatGPT-16k and GPT-4o~\cite{gpt4o}. For each model, we conduct tests using two types of prompts. The first type directly describes the task by providing an instruction text $\mathbf{y}$ and accompanying it with the text $\mathbf{x}$. The second type uses the Chain-of-Thought (CoT) prompting~\cite{wei2022chain}, incorporating the phrase ``let's think step by step'' into the instruction text. See more details in Appendix~\ref{sec:app:detail_benchmark}.

\paragraph{Evaluation Metric}
As the generated cell content in this task consists of numerical values, we utilize commonly employed metrics in regression tasks, namely the Root Mean Square Error (RMSE).
$$RMSE=\sqrt{\frac{\sum_{i=1}^n\left(y_i-\tilde{y}_i \right)^2}{n}}$$
where $n$ represents the number of cells, $y_i$ and $\tilde{y}_i$ represent the contents of the cell at index $i$ (for 2D tables, the index is calculated after reshaping to the 1D table) in the ground truth table and the output table, respectively.

We also report the Error Rate (ER) for each cell, defining a cell as erroneous if its content does not exactly match the ground truth. 
\paragraph{Grouping by Event Difficulty}
Furthermore, we categorize the eight types of events into three groups based on assessed difficulty: \texttt{Goals}, due to direct descriptions of scores in the original text, and \texttt{Red Cards}, due to their rare occurrence are categorized into the \emph{Easy} section. \texttt{Shots} and \texttt{Fouls}, due to their varied expressions and descriptions, are classified into the \emph{Hard} section. The remaining four event types are classified as \emph{Medium} section. We report the RMSE and ER for each model across different difficulty categories to provide a more comprehensive analysis.

\section{Experiments and Analyses}
In this section, we will benchmark the performance of current state-of-the-art LLMs on the \textsc{LiveSum} dataset, and further evaluate the effectiveness and generalization of our proposed approach. We aim to answer the following research questions:

\paragraph{\textsc{RQ1} (Benchmarking)} \ How do the current state-of-the-art LLMs perform on this dataset in fine-tuning and zero-shot settings?
\paragraph{RQ2 (Effectiveness)} \ \ \ How does our proposed \textsc{T}\textsuperscript{3} pipeline impact model performance?
\paragraph{RQ3 (Generalization)} How effective is the \textsc{T}\textsuperscript{3} pipeline when applied to other real-world datasets for the text-to-table generation task?

\begin{table*}[!t]

    \renewcommand\arraystretch{1.06}
\small
    \begin{center}           
    \begin{tabular}{m{3.9cm}|m{1.0cm}<{\centering}m{1.0cm}<{\centering}|m{1.0cm}<{\centering}m{1.0cm}<{\centering}|m{1.0cm}<{\centering}m{1.0cm}<{\centering}||m{1.0cm}<{\centering}m{1.0cm}<{\centering}}
            \toprule
            \multirow{2}{*}{\textbf{Model}} & \multicolumn{2}{c|}{\textbf{Easy}} &  \multicolumn{2}{c|}{\textbf{Medium}}  &\multicolumn{2}{c||}{\textbf{Hard}} & \multicolumn{2}{c}{\textbf{Average}}  \\
            & \scriptsize {RMSE} & \scriptsize  {ER} & \scriptsize {RMSE} & \scriptsize  {ER} & \scriptsize {RMSE} & \scriptsize  {ER} & \scriptsize {RMSE} & \scriptsize  {ER} \\
            \midrule
            \multicolumn{9}{l}{$\vardiamondsuit$ \emph{\textbf{Fine-Tune}}} \\
Mistral-7B-Instruct-v0.2  &\underline{1.045}& \textbf{38.36}& 3.832& 85.11& 7.115& 95.52& 4.564& 76.03\\ 
LLaMA-2-Chat 7B   & 1.047& \underline{38.41}& \underline{3.728}& \underline{84.91}& \underline{7.107}& \underline{95.40}& \underline{4.512}& \underline{75.91}\\ 
LLaMA-2-Chat 13B & \textbf{1.043} & 39.22 & \textbf{3.587} & \textbf{84.60} & \textbf{6.671} & \textbf{94.42} & \textbf{4.287} & \textbf{75.71}\\
\midrule
            \multicolumn{9}{l}{$\vardiamondsuit$ \emph{\textbf{Zero-Shot}}} \\
LLaMA-2-Chat 13B & 0.775 & 33.29 & 4.554 & 87.37 & 5.203 & 93.29 & 4.279 & 75.33 \\
LLaMA-2-Chat 13B \textsc{(CoT)} & 0.780 & 31.83 & 4.376 & 87.42 & 5.088 & 92.35 & 4.162 & 74.75 \\
LLaMA-2-Chat 70B & 0.410 & 12.34 & 3.189 & 88.59 & 4.941 & 92.41 & 3.455 & 70.48 \\
LLaMA-2-Chat 70B \textsc{(CoT)} & 0.450 & 12.86 & 3.221 & 89.25 & 5.314 & 94.24 & 3.613 & 71.40 \\
ChatGPT & 0.200&\ \  8.06& 2.864& 72.73& 4.257& 90.62& 3.008& 61.03\\ 
ChatGPT \textsc{(CoT)} & 0.229& 10.61& 2.809& 72.75& 4.087& 90.38& 2.911& 61.62\\ 
Claude 2.1 &1.014& 10.08& 2.581& 63.99& 4.621& 90.58& 3.171& 57.16\\ 
Claude 2.1 \textsc{(CoT)} &1.496& 14.06& 2.291& 61.70& 4.081& 90.38& 2.918& 56.96\\ 
Mistral Large &  \textbf{0.005}& \ \ \textbf{0.27}& 2.385& 52.45& \underline{2.712}& \underline{84.62}& 2.209& 47.45\\ 
Mistral Large \textsc{(CoT)} & \underline{0.018}& \ \ \underline{0.73}& 2.311& 51.82& \textbf{2.608}& \textbf{84.08}& \underline{2.139}& 47.12\\ 
GPT-4  & 0.156& \ \ 4.64& \textbf{1.167}& \underline{46.05}& 4.114& 88.53& 2.273& \underline{46.32}\\ 
GPT-4 \textsc{(CoT)} &0.154& \ \ 4.38&\underline{1.173}& \textbf{45.86}& 3.981& 88.73& 2.225& \textbf{46.20}\\ 
Claude 3 Opus & 0.078& \ \ 2.52& 1.617& 51.36& 3.713& 88.06& 2.253& 48.33\\ 
Claude 3 Opus \textsc{(CoT)} &0.040& \ \ 1.59& 1.642& 49.60& 3.265& 87.86& \textbf{2.079}& 47.17\\ 
\midrule
\multicolumn{9}{l}{$\vardiamondsuit$ \emph{\textbf{Zero-Shot with}} \textbf{T\textsuperscript{3}}} \\
Claude 2.1 \textsc{(T\textsuperscript{3})} & 0.193& \ \ 8.95& 1.965& 44.99& 2.751& 72.15& 2.066& 42.77\\ 
Mistral Large \textsc{(T\textsuperscript{3})}& 0.191& \ \ 8.82& 1.596& 42.37& 2.136& 69.23& 1.631& 40.70\\
GPT-4 \textsc{(T\textsuperscript{3})} & \textbf{0.056}& \ \ \textbf{3.18}& \underline{0.854}& \underline{25.83}& \underline{1.219}& \underline{46.22}& \underline{0.929}& \underline{25.27}\\ 
Claude 3 Opus \textsc{(T\textsuperscript{3})} & \underline{0.081}&\ \ \underline{5.30}& \textbf{0.406}& \textbf{14.79}& \textbf{0.477}& \textbf{21.29}& \textbf{0.438}& \textbf{14.04}\\ 
            \bottomrule
    \end{tabular}
    \end{center}
    \vspace{-0.15in}
\caption{The performance of various LLMs under three settings, showing RMSE and error rate across three difficulty categories and overall average. We \textbf{bold} the best results and \underline{underline} the second-best results in each setting.}
\vspace{-10pt}
    \label{tab:main_result}
\end{table*}

\begin{table}[!t]

\renewcommand\arraystretch{1.08}
\small
    \centering
    \begin{tabular}{m{1.17cm}m{1.1cm}<{\centering}m{1.1cm}<{\centering}m{1.1cm}<{\centering}m{1.1cm}<{\centering}}
    \toprule
     \multirow{2}{*}{\textbf{Model}} & \textbf{Easy} & \textbf{Medium} &\textbf{Hard} & \textbf{Average} \\
      & \scriptsize {RMSE/ER}& \scriptsize {RMSE/ER}& \scriptsize {RMSE/ER}& \scriptsize {RMSE/ER} \\
      \midrule
      \multicolumn{5}{l}{$\vardiamondsuit$ {\textbf{gpt-3.5-turbo-16k}}} \\
     0-shot                                                             & 0.22/12.0 & 1.76/65.1 & 3.22/\textbf{87.2} & 2.08/57.3  \\
      \scriptsize $\ \ \Diamond$ w/ \textsc{CoT}                        & 0.23/12.7 & \textbf{1.75}/\textbf{64.4} & \textbf{3.14}/87.4 & \textbf{2.05}/\textbf{57.2}  \\
      1-shot                                                            & 0.25/13.5 & 2.29/71.9 & 4.33/90.4 & 2.76/61.9  \\
      \scriptsize $\ \ \Diamond$ w/ \textsc{CoT}                        & \textbf{0.18}/\textbf{8.82} & 3.18/73.8 & 4.52/91.5 & 3.24/62.0  \\
    \midrule
    
      \multicolumn{5}{l}{$\vardiamondsuit$ {\textbf{gpt-4o}}} \\
     0-shot                                                             &\underline{0.00}/0.13 & 1.51/50.4 & \textbf{1.82}/78.5 & \underline{1.43}/44.8  \\
      \scriptsize $\ \ \Diamond$ w/ \textsc{CoT}                        & 0.00/0.20 & 1.31/\textbf{41.7} & \underline{2.00}/\underline{78.1} & \textbf{1.41}/\textbf{40.4}  \\
      1-shot                                                            & 0.00/0.19 & \underline{1.24}/49.0 & 2.19/78.8 & 1.44/44.3  \\
      \scriptsize $\ \ \Diamond$ w/ \textsc{CoT}                        & {0.00}/\underline{0.13} & \textbf{1.19}/\underline{45.0} & 2.72/84.3 & 1.63/\underline{43.6}  \\
      5-shots                                                           & 0.00/0.20 & 1.71/57.2 & 2.15/80.0 & 1.65/48.6  \\
      \scriptsize $\ \ \Diamond$ w/ \textsc{CoT}                        & \textbf{0.00}/\textbf{0.13} & 1.55/56.3 & 2.02/\textbf{72.9} & 1.53/46.4  \\
     \bottomrule
    \end{tabular}
    \vspace{-0.04in}
    \caption{The performance of two LLMs under the few-shot setting. We \textbf{bold} the best results and \underline{underline} the second-best results.}
    \vspace{-0.1in}
    \label{tab:few_shot}
\end{table}

\subsection{Benchmarking (RQ1)}

We first analyze the performance of existing state-of-the-art LLMs on the \textsc{LiveSum} dataset under fine-tuning, zero-shot and few-shot settings, with results displayed in Figure~\ref{fig:main_result}, Table~\ref{tab:main_result} and Table~\ref{tab:few_shot}. Overall, the performance of most models in the zero-shot setting far exceeds that in the fine-tuning setting, indicating that the previous state-of-the-art fine-tuning method has limited capability for information integration, and there is substantial room for improvement on this benchmark.
In the zero-shot setting, it is noteworthy that most models show a slight improvement in both metrics after applying \textsc{CoT}. Among them, the best-performing models are Mistral Large, GPT-4, and Claude 3 Opus, which are nearly comparable. They achieve RMSEs ranging from 2.08 to 2.27 and error rates between 46.20\% and 48.33\%. Nevertheless, this still highlights a notable deficiency in the information integration capabilities of LLMs in the zero-shot settings, underscoring the challenges and significance of our benchmark. The results in Table~\ref{tab:few_shot} indicate that few-shot learning does not consistently improve model performance on this task. However, there is a significant reduction in the error rate for the hard portion under the 5-shots-\textsc{CoT} setting. We speculate that in such complex tasks, the model may not effectively learn and understand the examples, and instead, be influenced by the examples, leading to incorrect answers. We then analyze performance across three categories of difficulty.

\paragraph{Easy Section}
It can be observed that the error rate of the fine-tuned models is generally around 40\%, with an RMSE close to 1. In the zero-shot setting, LLaMA-2-Chat, ChatGPT, and Claude 2.1 models exhibit relatively poor performance, occasionally producing anomalously large values. The error rates of the other models generally remain below 5\%, with RMSEs less than 0.2. Among these, the Mistral Large model performs the best, with both metrics significantly lower than other models.

\paragraph{Medium Section}
The medium section exhibits the greatest variation among models and serves as a crucial determinant of overall model performance. We organize the models based on their performance, with error rates in the zero-shot setting ranging from 89.25\% down to 45.86\%. GPT-4 with \textsc{CoT} performs the best, achieving the lowest error rate, which still indicates the suboptimal capabilities of LLMs.

\paragraph{Hard Section}
The charts clearly show that in the hard section, the zero-shot method shows minimal enhancement compared to fine-tuning, as most error rates are around 90\%. However, Mistral Large is an exception, achieving a lower error rate of 84.08\%, which demonstrates the challenging nature of the hard section. 



\subsection{Effectiveness (RQ2)}
\label{sec:ablation_study}

\begin{table}[!t]

\renewcommand\arraystretch{1.08}
\small
    \centering
    \begin{tabular}{m{1.15cm}m{1.1cm}<{\centering}m{1.1cm}<{\centering}m{1.1cm}<{\centering}m{1.1cm}<{\centering}}
    \toprule
     \multirow{2}{*}{\textbf{Model}} & \textbf{Easy} & \textbf{Medium} &\textbf{Hard} & \textbf{Average} \\
      & \scriptsize {RMSE/ER}& \scriptsize {RMSE/ER}& \scriptsize {RMSE/ER}& \scriptsize {RMSE/ER} \\
      \midrule
     GPT-4                                  & 0.16/4.6 & 1.17/46.1 & 4.11/88.5 & 2.27/46.3  \\
      w/ \textsc{CoT}                       & 0.15/4.4 & 1.17/45.9 & 3.98/88.7 & 2.23/46.2  \\
      w/ \textsc{T\textsuperscript{3}{m}}                     & \textbf{0.00}/\textbf{0.1} & 1.42/43.2 & 2.46/82.8 & 1.62/42.3  \\
      w/ \textsc{T\textsuperscript{3}{d}}           & 0.09/4.5 & \underline{1.12}/\underline{40.1} & \underline{2.23}/\underline{81.4} & \underline{1.42}/\underline{41.5}  \\
    \midrule
      w/ \textsc{T}\textsuperscript{3}      & \underline{0.06}/\underline{3.2} & \textbf{0.85}/\textbf{25.8} & \textbf{1.22}/\textbf{46.2} & \textbf{0.93}/\textbf{25.3} \\
     \bottomrule
    \end{tabular}
    \caption{The ablation study results comparing the performance of different prompting methods. We \textbf{bold} the best results and \underline{underline} the second-best results.}
    \vspace{-0.1in}
    \label{tab:ablation_study}
\end{table}

We apply the \textsc{T}\textsuperscript{3} pipeline to four LLMs: Claude 2.1, Mistral Large, GPT-4, and Claude 3 Opus; the rest are not applicable to this method due to their ineffective extraction of tuples from inputs of such lengths, resulting in a minimal number of tuples or a substantial duplication of the same tuple. The implementation details are discussed in  Appendix~\ref{sec:app:detail_t3}. From Figure~\ref{fig:main_result} and Table~\ref{tab:main_result}, it is observable that Claude 2.1 and Mistral Large exhibit similar improvements after applying the \textsc{T}\textsuperscript{3} method, both showing slight enhancements over the best results in the zero-shot setting, with reductions in RMSE of 34.9\% and 26.2\% respectively and decreases in error rate of 25.2\% and 14.2\%. In contrast, GPT-4 and Claude 3 Opus display substantial improvements after implementing the \textsc{T}\textsuperscript{3} method, with RMSE reductions of 59.1\% and 80.6\%, respectively, and error rate reductions of 45.4\% and 70.9\%. The reductions in these metrics are primarily reflected in the significant decrease in the error rate for the hard sections, creating a clear distinction from the zero-shot approaches. 
We then conduct an ablation study to evaluate the impact of the \textsc{T}\textsuperscript{3} method on model performance. We experiment with two variant methods: \textsc{T\textsuperscript{3}-Merged} \textsc{(T\textsuperscript{3}{m})} and \textsc{T\textsuperscript{3}-Direct-Execution} \textsc{(T\textsuperscript{3}d)}. The former method involves using a single prompt that instructs the model to first extract relevant tuples before generating the table, while the latter one modifies the second step of the \textsc{T}\textsuperscript{3} method to be directly executed by the LLM, rather than using code generation (see Appendix~\ref{sec:app:ablation_prompts} for more details). Table~\ref{tab:ablation_study} presents the results using GPT-4 as an example. It is apparent that using \textsc{T}\textsuperscript{3} considerably enhances the model's overall performance, leading significantly in overall metrics, with average reductions in RMSE and error rate of 59.1\% and 45.4\%, respectively. Compared to GPT-4 and \textsc{CoT}, the variants \textsc{T\textsuperscript{3}{m}} and \textsc{T\textsuperscript{3}{d}} also exhibit notable improvements, with respective reductions in RMSE of 28.7\% and 37.6\%, and error rates of 8.6\% and 10.3\%. Full ablation studies are in Appendix~\ref{sec:app:ablation_full}.

\subsection{Generalization (RQ3)}

To examine the generalization capabilities of \textsc{T}\textsuperscript{3},
we apply \textsc{T}\textsuperscript{3} to two additional datasets designed to test text-to-table performance and compare it with previous methods. Section~\ref{sec:struct_bench} involves table generation without the need for information integration, while Section~\ref{sec:struct_sum} focuses on table generation without instructions.

\subsubsection{Performance of \textsc{T}\textsuperscript{3} on \textsc{Struc-Bench} Table Dataset}
\label{sec:struct_bench}
\begin{table}[t]

    \renewcommand\arraystretch{1.06}
\footnotesize
    \centering

    \begin{tabular}{p{2.3cm}|p{1cm}<{\centering}p{1.1cm}<{\centering}|p{1.55cm}<{\centering}}
    \toprule
       \multirow{2}{*}{\textbf{Metric}} &  \multicolumn{2}{c|}{\textbf{Zero-Shot}}&\textbf{Fine-Tune}\\
       &   {ChatGPT}  &  w/ \textsc{T}\textsuperscript{3} & LLaMA-7B\\
       \midrule
{SacreBLEU}  & {77.58} & {\underline{78.91}} & \textbf{90.60} \\
{ROUGE-L}  & {86.11} & {\underline{88.36}}& \textbf{88.98}  \\
{BERTScore}  & {96.75} & {\underline{97.34}}& \textbf{98.54}  \\
{BLEURT}  & {64.66} & \textbf{{67.47}}& \underline{66.07}  \\
{BARTScore}  & \ {-2.08} & \ \underline{{-1.90}}& \ \textbf{-0.69}  \\
{Content P-Score}    & {\ \ 6.84}  &\ \  {\underline{7.29}}& \ \ \textbf{7.69}  \\
{Format P-Score} & \ \ \underline{9.70}   &\ \  {\textbf{9.88}}& \ \ {8.60}  \\
{Content H-Score}  & \ \ \underline{1.66} &\ \  {\textbf{1.68}}& \ \ {1.65}  \\
{Format H-Score}  & \ \ {3.28} & \ \ {\textbf{3.63}}& \ \ \underline{3.61}  \\
    \bottomrule
    \end{tabular}
    \caption{The evaluation results on the test set of \textsc{Struc-Bench} Table dataset with nine metrics. We \textbf{bold} the best results and \underline{underline} the second-best results.}
    \vspace{-0.1in}
    \label{tab:exp_strucbench}
\end{table}

We test the performance of \textsc{T}\textsuperscript{3} on the text-to-table benchmark \textsc{Struc-Bench} Table~\cite{tang2023strucbench}. This benchmark is based on the \textsc{Rotowire} dataset~\cite{wiseman2017challenges} and employs traditional evaluation metrics, prompting score(\textit{P-Score}), and heuristical score(\textit{H-Score}), to conduct a comprehensive assessment of the output tables. They also introduce a fine-tuning approach incorporating row and column header information in the training instructions. We argue that this comparison with the zero-shot method is problematic. 
In zero-shot settings, due to the absence of a fine-tuning process, the format of the output tables is uncertain. For example, in ground-truth tables, cells that are left blank may be filled with terms such as ``unknown'' or ``not mentioned'' by the model, substantially impacting similarity-based metrics. Hence we modify the model's outputs under zero-shot settings before reporting the results.
We evaluate the performance of ChatGPT with and without employing the \textsc{T}\textsuperscript{3} method and also compare it to the fine-tuned LLaMA-7B model proposed by \citet{tang2023strucbench}. Results are detailed in Table~\ref{tab:exp_strucbench}. It is observable that the application of the \textsc{T}\textsuperscript{3} method results in significant improvements across all metrics, with some measures outperforming the fine-tuned model. Further details on the experiments and analysis of the results are discussed in Appendix~\ref{sec:app:details_strucbench}.

\subsubsection{Performance of \textsc{T}\textsuperscript{3} on \textsc{Wiki40b} Dataset}
\label{sec:struct_sum}

We intend to evaluate the performance of our proposed approach in the text-to-table task without instructions and ground-truth tables. In line with the pioneering work of \textsc{StructSum}~\cite{jain2024structsum}, we experiment on a randomly sampled set from the English section of \textsc{Wiki40b} dataset~\cite{49029wiki40B}. As there is no ground-truth table for the text in the dataset, they leverage LLMs and propose \textsc{Auto-QA} Coverage as an evaluation metric:
\begin{equation*}
\text{Cov}(\mathcal{T})=\frac{\sum_{i=1}^{|G(\mathcal{S})|}E_{(q_i,a_i)}[Q(\mathcal{T},q_i)]}{|G(\mathcal{S})|}
\end{equation*}
where $G(\mathcal{S})$ is the list of Question-Answer pairs $(q_i,a_i)$ generated by the LLM based on text $\mathcal{S}$, $Q(\mathcal{T},q)$ is the LLM's answer to question $q$ based on table $\mathcal{T}$, and $E_{(q,a)}[x]$ is the LLM's evaluation of whether answer $a$ and $x$ are equivalent for question $q$.
On top of this evaluation, we add a step where an LLM is used to pre-screen each $(q,a)$ pair based on text $\mathcal{S}$, filtering out any pairs where the question can not be correctly answered. This process further assures the quality of the QA pairs generated by $G$. We opt for ChatGPT as the LLM for evaluation and randomly sample 500 passages for the test dataset following \citet{jain2024structsum}. We also introduce \textsc{T}\textsuperscript{2}(\textbf{T}ext-\textbf{T}uple) which treats the intermediary tuples from \textsc{T}\textsuperscript{3} as a single table $\mathcal{T}$. We aim to investigate the extent of information loss during the conversion from tuples to tables through this configuration. Figure~\ref{fig:auto_qa_cover} shows the \textsc{Auto-QA} Coverage of three methods. The curve indicates the percentage of generated tables meeting a given coverage threshold. Overall, T\textsuperscript{3} demonstrates a substantial improvement over the prior Divide-and-Generate method~\cite{jain2024structsum}. For example, when the coverage threshold is set to 70\%, about 65\% of data reach this threshold after applying T\textsuperscript{3}, compared to only 50\% with the preceding approach. It is important to note that \textsc{T}\textsuperscript{2} outperforms \textsc{T}\textsuperscript{3}, with 83\% of data retaining the same coverage after tuple extraction. This suggests that information loss occurs during the transformation of raw tuples into structured tables. Exploring ways to mitigate this loss represents an essential area for further research. More details are discussed in Appendix~\ref{sec:app:details_wiki40b}.

\begin{figure}[t]
  \centering
    \begin{tikzpicture}[scale=0.90]
    \begin{axis}[ylabel={\textsc{Auto-QA} Coverage (\%)},
    ylabel style={yshift=3.0em},
    xlabel={Percentage (\%)},tick pos=left,
    legend pos=north east,
    legend style={at={(0.5,1.05)},anchor=west,
        anchor=south,
        legend columns=1, 
        },
        xtick={0,10,20,30,40,50,60,70,80,90,100},ytick={10,20,30,40,50,60,70,80,90,100},
    minor xtick={0,10,20,30,40,50,60,70,80,90,100},minor ytick={10,20,30,40,50,60,70,80,90,100},grid=minor,xmin=-5,xmax=105,ymin=10,ymax=105,y label style={at={(axis description cs:0.08,.5)},anchor=south},legend cell align={left},
        nodes near coords=,axis equal image]
    \addplot[line width=1.6pt,mark size=1.2pt,mark=*,cRed] table[x index=0,y index=1,col sep=comma] {dataplot.dat};
    \addlegendentry{\small Divide-and-Generate~\cite{jain2024structsum} }
    \addplot[line width=1.6pt,mark size=1.2pt,mark=*,cBlue] table[x index=0,y index=2,col sep=comma] {dataplot.dat};
    \addlegendentry{\small \textsc{T}\textsuperscript{3}(Text-Tuple-Table)\ }
    \addplot[line width=1.6pt,mark size=1.2pt,opacity=0.5,mark=*,cGray] table[x index=0,y index=3,col sep=comma] {dataplot.dat};
    \addlegendentry{\small \textsc{T}\textsuperscript{2}(Text-Tuple)}
    \end{axis}
    \end{tikzpicture}
    \caption{\textsc{Auto-QA} coverage of the three methods. The point $(P,C)$ means $P\%$ of the data can achieve a coverage of $C\%$ or higher measured using \textsc{Auto-QA}.}
    \vspace{-0.2in}
    \label{fig:auto_qa_cover}
\end{figure}

\subsection{Case Studies}
We present specific case studies on the outputs of different models on the \textsc{LiveSum} dataset in Appendix~\ref{sec:app:case_study}. These cases directly demonstrate the effectiveness of our proposed method. Additionally, we summarize some common errors of the model after applying \textsc{T}\textsuperscript{3} and areas for improvement.
\section{Related Work}

\paragraph{Text-to-Table Generation}

Many studies have been proposed to perform text-to-table generation, converting it into sequence-to-sequence problems~\cite{wu2022text,li-etal-2023-sequence-sequence}, or framing them as question-answering problems~\cite{sundar2024gtbls}. With the rise of LLMs, some research has also explored evaluating LLMs under fine-tuning or zero-shot settings, and it shows that fine-tuning yields highly effective results~\cite{tang2023strucbench,sundar2024gtbls}. However, these methods employ datasets that only require the model to extract relevant information from text and populate tables, which significantly limits the scope of this task. Therefore, we introduce a new challenging dataset and propose a universal solution that greatly enhances the performance of LLMs under zero-shot setting.

\paragraph{LLMs for Information Extraction}

Information Extraction (IE) is critical and foundational for many downstream tasks in NLP. Many works have been conducted to leverage LLMs and provide effective solutions for IE tasks within a generative framework~\cite{ma-etal-2023-large, lu-etal-2023-event, wan-etal-2023-gpt,zhou2024universalner}. Recent progress in LLMs also has led to the development of unified frameworks that model various IE tasks and domains~\cite{wang2023instructuie,sainz2024gollie}. This aligns with our intention to harness this capability to address general text-to-table generation tasks.

\paragraph{LLM Promptings}

Prompt engineering has been essential for enhancing LLMs and has demonstrated great success across a wide range of applications~\cite{wei2022chain,dua-etal-2022-successive,li2023structured,DBLP:conf/acl/ChanLCLSWS23,wang2024chainoftable,DBLP:journals/corr/abs-2408-02559, DBLP:conf/ijcnlp/ChanLCCSWS23, lin2024constrained, DBLP:conf/eacl/ChanCWJFLS24}. Among the various prompting techniques, we find the decomposed prompting~\cite{khot2022decomposed}, which breaks down complex tasks into easier sub-tasks via prompting, highly effective for text-to-table generation. \citet{jain2024structsum} adopts this idea by breaking the text into small pieces for table generation. However, we argue that such a decomposition approach is impractical because text is not always easily divisible. For example, in our dataset, such division might result in adjacent sections describing the same event, causing errors. Hence we propose a more intuitive and broadly applicable task decomposition pipeline.

\section{Conclusion}

In this work, we introduce \textsc{LiveSum}, a novel and challenging benchmark dataset for assessing 
 the capability of models to integrate information in the text-to-table generation, along with a robust pipeline named  \textsc{T}\textsuperscript{3}. Experimental results show that current LLMs underperform on our dataset in both fine-tuning and zero-shot settings; however, significant improvements are observed after applying our proposed \textsc{T}\textsuperscript{3} pipeline. Our method can also be applied to any text-to-table dataset, enabling LLMs to outperform previous methods in zero-shot settings.

\section*{Limitations}
Although our \textsc{LiveSum} benchmark extensively evaluates the information integration capabilities of LLMs, we have not yet tested their performance in a few-shot setting. Despite the challenge posed by the token length of live commentary for few-shot settings, we reserve this aspect for future work. Furthermore, while our proposed \textsc{T}\textsuperscript{3} pipeline significantly improves performance on several state-of-the-art LLMs, it cannot be effectively applied to LLMs that are deficient in tuple extraction capabilities, as it fails in the first stage and cannot proceed to the next phase. Developing methods that boost performance on such LLMs remains a valuable area for future research.

\section*{Ethics Statement}
When constructing the \textsc{LiveSum} dataset, we sample texts of live football match commentary from the open-access BBC Sports official website. We apply LLMs to paraphrase this live commentary and conduct manual reviews to ensure no harmful content is generated. We also anonymize the data using named entity recognition (NER) technology combined with player rosters. The datasets used in our experiments, \textsc{Struc-Bench}~\cite{tang2023strucbench} and \textsc{Wiki40b}~\cite{49029wiki40B}, are open-source, and all experiments adhere to their intended use for research purposes. Therefore, to the best of the authors' knowledge, we believe that this work introduces no additional risk.

\section*{Acknowledgements}
The authors of this paper were supported by the NSFC Fund (U20B2053) from the NSFC of China, the RIF (R6020-19 and R6021-20), and the GRF (16211520 and 16205322) from RGC of Hong Kong.

\bibliography{texttupletable}

\appendix
\clearpage
\newpage
\begin{center}
    {\Large\textbf{Appendices}}
\end{center}

\section{Details of \textsc{LiveSum} Dataset}
\label{sec:app:construct_deil}

This section extends Section~\ref{sec:dataset}, where we discuss additional details of dataset construction, and present detailed statistical information and a specific example of the dataset.

\subsection{Live Commentary Generation}

\paragraph{Paraphrasing}
We first utilize ChatGPT to paraphrase the original text to make it closer to the language used by the commentator while ensuring a certain degree of diversity. The prompting template used in this step is as follows:
\vspace{-0.05in}
\begin{tcolorbox}[title={Paraphrase Prompting Template}, colback = cBlue_1!10, colframe = cBlue_9,  coltitle=white,fonttitle=\bfseries\small, center title,fontupper=\small,fontlower=\small]
You are a football commentator, paraphrase the sentence and don't make it too long: \texttt{<Original Text>}
\end{tcolorbox}
\vspace{-0.05in}
\noindent where \texttt{<Original Text>} represents the original live commentary on the webpage. After the transcription of each live commentary segment, we manually inspect whether the meaning of each sentence has been altered. If any issues are identified, we make manual revisions accordingly.

\paragraph{Data Anonymization}
We then perform the data anonymization. First, we obtain the list of players participating in each match from the BBC Sports official website. Then, using a combination of string matching and NER techniques~\cite{qi2020stanza}, we match fully detected names to players on the list based on textual similarity. Each individual is assigned a unique number and recorded in the format \texttt{Player<number>(<team>)}, where \texttt{<number>} is a positive integer and \texttt{<team>} is the team that player belongs to. We anonymize the team names as ``Home Team'' and ``Away Team''. The inclusion of team names is intended to eliminate the need for the model to infer the team affiliation of each player, thus avoiding any potential interference with our evaluation of their information integration ability. We also manually inspect whether there is any missed anonymization. If any are found, we make manual revisions accordingly.

\subsection{Summary Table Generation}

\paragraph{Annotation}
We recruit five workers who voluntarily agreed to participate in this research without receiving any compensation to annotate the events and employ a majority voting approach to resolve any disagreements. These five workers are all postgraduate students who have 
. We randomly sample 100 cases with discrepancies and find that all errors are due to carelessness. Therefore, we conclude that the quality of the annotation quality is assured.

\paragraph{Integration}
After manual annotation and review, the data are aggregated into several tuples, and we count the occurrence of each tuple to obtain the integrated result. Finally, we sequentially fill in the column headers of the table with the eight events, convert the team names to  \texttt{Home Team} and \texttt{Away Team}  for the row headers of the table, and then fill in the corresponding counts into the cells to generate the final summary table.

\subsection{Statistics}

Table~\ref{tab:statistics} presents the statistical data regarding text length and the frequency of various events within the train and test sets of the \textsc{LiveSum} dataset. It can be seen that the statistical metrics across the train set and test set are comparably uniform, consistent with our method of completely random division.

\begin{table}[!t]
\small
    \renewcommand\arraystretch{1.08}
    \centering
    \begin{tabular}{p{2.5cm}p{1.5cm}<{\centering}p{1.5cm}<{\centering}}
    \toprule
     &  \textbf{Train Set} & \textbf{Test Set} \\
     \midrule
     \# Instances & 3,017 & \ \ \ 754 \\
     \midrule
     \multicolumn{2}{l}{\emph{Average Text Length}} \\
     Words &  1,258 & 1,250  \\
     Chars &  6,852 & 6,816  \\
     \midrule
     \multicolumn{2}{l}{\emph{Average Event Occurrence Frequency}} \\
     Goals &  \ \ 1.38 & \ \ 1.39   \\
     Shots &   12.71 &  12.55 \\
     Fouls &   10.60 &  10.57 \\
     Yellow Cards & \ \ 1.74 &  \ \ 1.76    \\
     Red Cards &  \ \ 0.04 & \ \ 0.03  \\
     Corner Kicks  & \ \ 5.25 & \ \ 5.24   \\
     Free Kicks&  10.34 & 10.30   \\
     Offsides&  \ \ 1.86 &   \ \ 1.84  \\
     \bottomrule
    \end{tabular}
    \vspace{-0.1in}
    \caption{Average statistics in \textsc{LiveSum} dataset.}
    \vspace{-0.2in}
    \label{tab:statistics}
\end{table}

\subsection{Example}

Figure~\ref{fig:full_example} presents an example of the live commentary and a summary table generated through the pipeline depicted in Figure~\ref{fig:construct_pipeline}.

\begin{figure*}[t]

    \footnotesize

\begin{tcolorbox}[colback = cBlue_1!5, colframe = cBlue_6,  coltitle=white,fonttitle=\bfseries\small,fontupper=\scriptsize,fontlower=\scriptsize]
\begin{small}
\begin{center}
\textbf{Textual Live Commentary}
\end{center}
\end{small}
      \texttt{Players are being announced for the lineup and getting ready for the game. The game is now underway with the start of the first half. Offside called on Home Team as Player5(Home Team) attempts a through ball to Player11(Home Team), who is caught offside. Player28(Away Team) makes a goal with a right-footed shot from the center of the box, giving the Away Team a 1-0 lead over the Home Team. Player8(Home Team) earns a free kick on the right side of the field. Player23(Away Team) commits a foul. Player9(Home Team) commits a foul. Player27(Away Team) earns a free kick in their own half. The Home Team earns a corner kick. Player5(Home Team)'s left footed shot from outside the box is saved in the bottom left corner after an assist from Player7(Home Team). Player26(Away Team) scores with a right-footed shot from outside the box, assisted by Player29(Away Team), Home Team 0, Away Team 2. The Home Team wins a corner kick. Player8(Home Team) commits a foul. Player21(Away Team) earns a free kick in their own half. Player29(Away Team) scores with a right-footed shot from the right side of the box, assisted by Player28(Away Team), Home Team 0, Away Team 3. Player7(Home Team) of the Home Team attempts a through ball, but Player11(Home Team) is flagged for being offside. Player29(Away Team) misses the goal with a high right footed shot from outside the box, assisted by Player26(Away Team). The Home Team wins a corner kick. Player10(Home Team) attempts a through ball, but Player11(Home Team) is offside for the Home Team. The Away Team wins a corner kick. Player5(Home Team) earns a free kick in their own half. Player21(Away Team) commits a foul. Player25(Away Team) fouls Player5(Home Team), who earns a free kick on the left wing. The Home Team wins a corner kick. Player10(Home Team) of the Home Team is caught offside after Player8(Home Team) attempts a through ball. At the end of the first half, the Home Team is trailing with a score of 0-3 against the Away Team. And we're back for the second half with the Home Team trailing 0-3 against the Away Team. Player2(Home Team) earns a free kick in their own half. Player23(Away Team) commits a foul. Player18(Home Team)'s header from the center of the box is saved in the bottom left corner. Player4(Home Team) commits a foul. Player29(Away Team) earns a free kick in their own half. Player4(Home Team) has received a yellow card for a reckless foul. Player29(Away Team) is currently delayed in the match due to an injury. The delay is finished and they are prepared to resume play. Player8(Home Team) is holding up the game due to an injury. The delay is finished and they are prepared to resume play. Player14(Home Team) commits a foul. Player23(Away Team) earns a free kick in the opponent's half. Player28(Away Team) missed the target with a shot from the right side of the box, with an assist from Player27(Away Team) after a quick counterattack. Player28(Away Team)'s right footed shot from the centre of the box was close, but missed to the right and then blocked. Player29(Away Team)'s shot from the center of the box is saved in the center of the goal after an assist from Player28(Away Team). The Away Team earns a corner kick. Player28(Away Team)'s shot from the center of the box is blocked with the assistance of Player23(Away Team). Player26(Away Team) attempts a through ball, but Player23(Away Team) is offside for the Away Team. Player14(Home Team) from the Home Team attempts a through ball, but Player11(Home Team) is flagged for being offside. The Home Team wins a corner kick. Player6(Home Team)'s close-range attempt is saved in the bottom left corner with an assist from Player3(Home Team)'s headed pass. Player18(Home Team) earns a free kick on the left wing after being fouled by Player20(Away Team). Player26(Away Team) attempts a through ball but Player20(Away Team) is offside for the Away Team. Player4(Home Team) commits a foul. Player28(Away Team) earns a free kick in their own half. Player27(Away Team)'s shot from the center of the box was blocked with the help of Player26(Away Team). Player6(Home Team) is being delayed in the match due to an injury. The delay is finished and they are prepared to resume play. Player18(Home Team)'s shot from outside the box with their right foot was too high, assisted by Player5(Home Team). Player6(Home Team) commits a foul on Player23(Away Team), resulting in a free kick being awarded on the left wing. The Away Team gets a corner kick. Player23(Away Team)'s left footed shot from outside the box following a corner is too high. Player25(Away Team) attempted a shot with his right foot from the right side of the box but it was too high, with an assist from Player28(Away Team). Player7(Home Team) earns a free kick in the opponent's half. Player36(Away Team) commits a foul. Player3(Home Team) misses the header to the left from the center of the box, assisted by Player16(Home Team) with a cross. Player28(Away Team) misses the target with a left-footed shot from long range. Player16(Home Team) attempts a through ball, but Player18(Home Team) is offside for the Home Team. The Home Team finishes the second half with a score of 0, while the Away Team has 3. Final score, Home Team 0, Away Team 3.} 

\tcblower

\begin{small}
\begin{center}
\textbf{Summary Table}
\end{center}
\end{small}
    \begin{tabular}{m{1.5cm}<{\centering}m{1.2cm}<{\centering}m{1.2cm}<{\centering}m{1.2cm}<{\centering}m{1.2cm}<{\centering}m{1.2cm}<{\centering}m{1.2cm}<{\centering}m{1.2cm}<{\centering}m{1.2cm}<{\centering}}
    \toprule
        \textbf{Team} &  \textbf{Goals} & \textbf{Shots} & \textbf{Fouls} & \textbf{Yellow Cards} & \textbf{Red Cards} & \textbf{Corner Kicks} & \textbf{Free Kicks} & \textbf{Offsides} \\
        \midrule
         Away Team & 3 & 12 & 6 & 0 & 0 & 3 & 6 & 2 \\
         \midrule
         Home Team & 0 & 5 & 6 & 1 & 0 & 5 & 6 & 6 \\
         \bottomrule
    \end{tabular}
\end{tcolorbox}
    \caption{An instance within the \textsc{LiveSum} dataset. The above section presents a complete textual live commentary that has undergone paraphrasing and data anonymization, while the table below represents a manually annotated summary table.}
    \label{fig:full_example}
\end{figure*}

\section{Implementation of \textsc{T}\textsuperscript{3} on \textsc{LiveSum} Dataset}
\label{sec:app:detail_t3}

This section is an extension of Section~\ref{sec:t3}, where we provide description of the implementation details of the \textsc{T}\textsuperscript{3} method applied to the \textsc{LiveSum} dataset.

\subsection{Instrction on \textsc{LiveSum} Dataset}
\label{sec:app:t3_instruction}
We present the instruction directives common to all experiments conducted on the \textsc{LiveSum} dataset, which are some rules related to this task:

\vspace{-0.05in}
\begin{tcolorbox}[title={Instruction on \textsc{LiveSum} Dataset}, colback = cBlue_1!10, colframe = cBlue_9,  coltitle=white,fonttitle=\bfseries\small, center title,fontupper=\small,fontlower=\small]
According to the live text, please count the number of: 1.goals, 2. shots, 3.fouls, 4.yellow cards, 5.red cards, 6.corner kicks, 7.free kicks, and 8.offsides for each team. Note that goals and saved attempts and blocked attempts and missed attempts are considered shots. Handball and dangerous play are also considered foul. The second yellow card is also considered a red card. Penalty is also considered as free kicks. 
\end{tcolorbox}
\vspace{-0.05in}

\subsection{Text-to-Tuple Prompts}
\label{sec:app:t3_s1}

Based on the characteristics of the text in the Livesum dataset, we have defined the format of tuples as  (\textit{player name}, \textit{team name}, \textit{event}) or (\textit{team name}, \textit{event}). Subsequently, we define the following prompting template:

\vspace{-0.05in}
\begin{tcolorbox}[title={\textsc{T}\textsuperscript{3} \\ Text-to-Tuple Prompting Template}, colback = cBlue_1!10, colframe = cBlue_6,  coltitle=white,fonttitle=\bfseries\small, center title,fontupper=\small]
\texttt{<Instruction>}

Please extract all the relevant event from the following passage, output them in (\textit{player name}, \textit{team name}, \textit{event}) or (\textit{team name}, \textit{event}) format. Constrain the event names to only the following options: 1.goals, 2.shots, 3.fouls, 4.yellow cards, 5.red cards, 6.corner kicks, 7.free kicks, and 8.offsides: 

\texttt{<Text>}
\end{tcolorbox}
\vspace{-0.05in}

\noindent where \texttt{<Instruction>} represents the text provided in Appendix~\ref{sec:app:t3_instruction}, and \texttt{<text>} is the live commentary text.

\subsection{Information Integration Prompts}
\label{sec:app:t3_s2}

As discussed in Section~\ref{sec:t3_s2}, \textsc{T}\textsuperscript{3} employs code generation as the default approach for information integration. The following is the prompting template we define, and Figure~\ref{fig:code_generated_by_gpt4} presents an example of Python code generated by GPT-4.

\vspace{-0.05in}
\begin{tcolorbox}[title={\textsc{T}\textsuperscript{3} \\ Information Integration Prompting Template}, colback = cBlue_1!10, colframe = cBlue_7,  coltitle=white,fonttitle=\bfseries\small, center title,fontupper=\small,fontlower=\small]

\textbf{Code Generation:}

\texttt{<Instruction>}

Please develop a Python code to consolidate these tuples as specified: 

\texttt{<Tuples>}
\end{tcolorbox}
\vspace{-0.05in}

\noindent where \texttt{<Instruction>} represents the text provided in Appendix~\ref{sec:app:t3_instruction}, and \texttt{<Tuples>} is the tuples extracted in the previous step.

\begin{figure*}[!t]
\begin{lstlisting}[language=Python, 
                   backgroundcolor=\color{cBlue_1!5}, 
                   frame=single, 
                   frameround=tttt, 
                   basicstyle=\ttfamily\scriptsize, 
                   commentstyle=\color{green!60!black}, 
                   keywordstyle=\color{blue}, 
                   stringstyle=\color{red!60!black}]
# Initialize a dictionary to store the counts
team_stats = {}

# List of tuples
tuples = [('player1', 'home team', 'goal'),
          ('player2', 'home team', 'shot'),
          ('player3', 'away team', 'foul')]

# Loop through the tuples
for player, team, event in tuples:
    # If the team is not in the dictionary, add it
    if team not in team_stats:
        team_stats[team] = {'goals': 0, 'shots': 0, 'fouls': 0, 'yellow cards': 0, 'red cards': 0, 
                            'corner kicks': 0, 'free kicks': 0, 'offsides': 0}
    
    # Update the counts based on the event
    if event == 'goal':
        team_stats[team]['goals'] += 1
        team_stats[team]['shots'] += 1
    elif event in ['shot', 'saved attempt', 'blocked attempt', 'missed attempt']:
        team_stats[team]['shots'] += 1
    elif event in ['foul', 'handball', 'dangerous play']:
        team_stats[team]['fouls'] += 1
    elif event == 'yellow card':
        team_stats[team]['yellow cards'] += 1
    elif event in ['red card', 'second yellow card']:
        team_stats[team]['red cards'] += 1
        if event == 'second yellow card':
            team_stats[team]['yellow cards'] += 1
    elif event == 'corner kick':
        team_stats[team]['corner kicks'] += 1
    elif event in ['free kick', 'penalty']:
        team_stats[team]['free kicks'] += 1
    elif event == 'offside':
        team_stats[team]['offsides'] += 1

# Print the counts for each team
for team, stats in team_stats.items():
    print(f"{team}: {stats}")
\end{lstlisting}
\caption{Python code generated by GPT-4 in response to the prompting as described in Appendix~\ref{sec:app:t3_s2}.}
\label{fig:code_generated_by_gpt4}
\end{figure*}

\subsection{Tuple-to-Table Prompts}
\label{sec:app:t3_s3}
In this step, we present the format of the generated table and specify the following prompting template:

\vspace{-0.05in}
\begin{tcolorbox}[title={\textsc{T}\textsuperscript{3} \\ Tuple-to-Table Prompting Template}, colback = cBlue_1!10, colframe = cBlue_8,  coltitle=white,fonttitle=\bfseries\small, center title,fontupper=\small]

\texttt{<Instruction>}

Please only output a table with the team name in CSV format with 2 rows based on the following tuples:

\texttt{<Tuples>}
\end{tcolorbox}
\vspace{-0.05in}

\noindent where \texttt{<Instruction>} represents the text provided in Appendix~\ref{sec:app:t3_instruction}, and \texttt{<Tuples>} is the tuples integrated by the code in the previous step.

\subsection{Prompts of Variant Methods}
\label{sec:app:ablation_prompts}
In Section~\ref{sec:ablation_study}, two variants of the \textsc{T}\textsuperscript{3} method are mentioned during the design of the ablation study. Here, we present the implementation details of these variants.

\paragraph{\textsc{T\textsuperscript{3}-Merged}}
This method combines all three stages of \textsc{T\textsuperscript{3}} into one, resulting in a single prompt:

\vspace{-0.05in}
\begin{tcolorbox}[title={\textsc{T\textsuperscript{3}-Merged} \\ Prompting Template}, colback = cBlue_1!10, colframe = cBlue_9!70!black,  coltitle=white,fonttitle=\bfseries\small, center title,fontupper=\small]

\texttt{<Instruction>}

Let's do the following things:

1. Extract all the relevant events from the following passage in (\textit{player name}, \textit{team name}, \textit{event}) or (\textit{team name}, \textit{event}) format.

2. Integrate these tuples.
                
3. Output a table with 2 rows in CSV format.

\texttt{<Text>}
\end{tcolorbox}
\vspace{-0.05in}
\noindent where \texttt{<Instruction>} represents the text provided in Appendix~\ref{sec:app:t3_instruction}, and \texttt{<text>} is the live commentary text.

\paragraph{\textsc{T\textsuperscript{3}-Direct-Execution}}
This method replaces code generation with direct execution in the second stage of \textsc{T}\textsuperscript{3}. The prompts are as follows:

\vspace{-0.05in}
\begin{tcolorbox}[title={\textsc{T\textsuperscript{3}-Direct} \\ Information Integration Prompting Template }, colback = cBlue_1!10, colframe = cBlue_9!70!black,  coltitle=white,fonttitle=\bfseries\small, center title,fontupper=\small]

\texttt{<Instruction>}

Please count all the information required and integrate these tuples:

\texttt{<Tuples>}
\end{tcolorbox}
\vspace{-0.05in}

\noindent where \texttt{<Instruction>} represents the text provided in Appendix~\ref{sec:app:t3_instruction}, and \texttt{<Tuples>} is the tuples extracted in the previous step. The first and third steps remain unchanged and are described exactly as in Appendix~\ref{sec:app:t3_s1} and ~\ref{sec:app:t3_s3}.

\section{Details of Benchmark Configuration}
\label{sec:app:detail_benchmark}

This section serves as an extension of Section~\ref{sec:experiment_setup}, providing additional details regarding the benchmark configuration.

\subsection{Fine-Tune Setting}
We use the open-source library named LLaMA-Factory\footnote{\href{https://github.com/hiyouga/LLaMA-Factory}{https://github.com/hiyouga/LLaMA-Factory}}~\cite{DBLP:journals/corr/abs-2403-13372} to fine-tune all models.
LoRA~\cite{DBLP:conf/iclr/HuSWALWWC22} is used as the fine-tuning paradigm to accommodate our computational resources.
The pre-trained weights are downloaded from the huggingface library~\cite{DBLP:conf/emnlp/WolfDSCDMCRLFDS20}.
We load the models with FP16 as the precision and optimize them with an Adam optimizer~\cite{DBLP:journals/corr/KingmaB14}.
The learning rate is set to 5e-5 and the batch size is 32. 
The maximum length for the input and generated sentence concatenation is 3,500. 
We warm up the model with 3,000 steps and evaluate the model every 300 steps. 
A linear scheduler is also used. 
The LoRA rank is set to 8, and the $\alpha$ is set to 32.

\subsection{Zero-Shot Setting}
The zero-shot and chain-of-thought inference for LLaMA-2 models are conducted on eight V100 GPUs~\cite{EcomScript}. 
When obtaining outputs from LLMs via APIs, we ensure deterministic results by setting the temperature to zero. 
In Table~\ref{tab:app:model_names}, we list the model names corresponding to different model types. Note that unless specifically stated otherwise, when a model type from the table is used, the model name corresponding to it in the table is the one being invoked.

\begin{table}[h]
\small
    \renewcommand\arraystretch{1.06}
    \centering
    \begin{tabular}{p{2.8cm}p{4cm}}
    \toprule
    \textbf{Model Type} & \textbf{Model Name} \\
    \midrule
Mistral Large & \texttt{mistral-large-2402} \\
ChatGPT & \texttt{gpt-3.5-turbo-0125} \\
GPT-4 & \texttt{gpt-4-0613} \\
Claude 2.1 & \texttt{claude-2.1} \\ 
Claude 3 Opus & \texttt{claude-3-opus-20240229} \\ 
    \bottomrule
    \end{tabular}
    \caption{Model types and their corresponding names of LLMs used in our experiments under zero-shot setting.}
    \vspace{-0.2in}
    \label{tab:app:model_names}
\end{table}

\subsection{Few-Shot Setting}

 Because the input lengths of \textsc{LiveSum} dataset generally exceed 2,000 tokens, even testing with 1-shot would surpass the maximum token limit (4,096) of most LLMs. Therefore, we select two models, ChatGPT-16k~\cite{openai2022chatgpt} and GPT-4o~\cite{gpt4o}, for supplementary experiments Regarding the selection of few-shot examples, we randomly choose samples from the training set where all events occur. In the Chain of Thought setting, the reasoning process is included in the examples.
 
\begin{table}[h]
\small
    \renewcommand\arraystretch{1.06}
    \centering
    \begin{tabular}{p{2.8cm}p{4cm}}
    \toprule
    \textbf{Model Type} & \textbf{Model Name} \\
    \midrule
ChatGPT-16k & \texttt{gpt-3.5-turbo-16k-0613} \\
GPT-4o & \texttt{gpt-4o-2024-05-13} \\
    \bottomrule
    \end{tabular}
    \caption{Model types and their corresponding names of LLMs used in our experiments under few-shot setting.}
    \vspace{-0.2in}
    \label{tab:app:model_names_2}
\end{table}

\subsection{Prompts}

Below are the two prompting templates we used to evaluate the baseline models:

\begin{tcolorbox}[title={Baseline Prompting Template}, colback = cBlue_1!10, colframe = cBlue_9,  coltitle=white,fonttitle=\bfseries\small, center title,fontupper=\small,fontlower=\small]

\textbf{Prompt w/o \textsc{CoT}}

\texttt{<Instruction>}

Please output a table with 2 rows in CSV format according to the following live text:

\texttt{<Text>}

\end{tcolorbox}
\vspace{-0.05in}
\begin{tcolorbox}[title={Baseline Prompting Template}, colback = cBlue_1!10, colframe = cBlue_9,  coltitle=white,fonttitle=\bfseries\small, center title,fontupper=\small,fontlower=\small]

\textbf{Prompt w/ \textsc{CoT}}

\texttt{<Instruction>}

Let's think step by step! At last, please output a table with 2 rows in CSV format according to the following live text:

\texttt{<Text>}

\end{tcolorbox}

\noindent where \texttt{<Instruction>} represents the text provided in Appendix~\ref{sec:app:t3_instruction}, and \texttt{<text>} is the live commentary text.

\subsection{Result Parsing}
As the table format in this experiment is fixed, most models can produce recognizable tables. We employ a comprehensive evaluation algorithm based on regular expressions to parse various types of output. For improperly formatted tables in the model's output, we filtered out those instances. This is because when the table format is disrupted, we are unable to obtain the meaning of each number in the output, and thus cannot calculate the metric. Moreover, this indicates that the model is even unable to perform the text-to-table generation task, which falls outside the scope of assessing its information integration abilities in the text-to-table process.

\section{Full Results of Ablation Study}
\label{sec:app:ablation_full}

\begin{table*}[!t]

    \renewcommand\arraystretch{1.1}
\small
    \begin{center}           
    \begin{tabular}{m{2.8cm}|m{1.0cm}<{\centering}m{0.9cm}<{\centering}|m{1.0cm}<{\centering}m{0.9cm}<{\centering}|m{1.0cm}<{\centering}m{0.9cm}<{\centering}||m{1.0cm}<{\centering}m{0.9cm}<{\centering}m{1.1cm}<{\centering}}
            \toprule
            \multirow{2}{*}{\textbf{Model}} & \multicolumn{2}{c|}{\textbf{Easy}} &  \multicolumn{2}{c|}{\textbf{Medium}}  &\multicolumn{2}{c||}{\textbf{Hard}} & \multicolumn{3}{c}{\textbf{Average}}  \\
            & \scriptsize {RMSE} & \scriptsize  {ER} & \scriptsize {RMSE} & \scriptsize  {ER} &\scriptsize {RMSE} & \scriptsize  {ER} &\scriptsize {RMSE} & \scriptsize  {ER}    & \scriptsize  {$\Delta_\text{ER}$} \\
            \midrule
Claude 2.1 & \underline{1.014}& \underline{10.08}& 2.581& 63.99& 4.621& 90.58 & 3.171& 57.16 & \\ 
Claude 2.1 \textsc{(CoT)} & {1.496}& 14.06& 2.291& \underline{61.70}& 4.081& 90.38 & 2.918& \underline{56.96} & $\downarrow$ \ \ 0.3\%\\ 
Claude 2.1 \textsc{(T\textsuperscript{3}{m})} & 1.144& 13.92& \underline{2.289}& 62.28& \underline{4.114}& \underline{90.12}& \underline{2.869}& 57.15 & $\downarrow$ \ \ 0.0\% \\ 
Claude 2.1 \textsc{(T\textsuperscript{3}{d})}  &{3.653}& 39.15& 2.444& 64.88& 5.621& 92.06 & 4.056& 65.24& $\uparrow$ 14.1\% \\ 
\rowcolor[gray]{0.9}
Claude 2.1 \textsc{(T\textsuperscript{3})}& \textbf{0.193}& \ \ \textbf{8.95}& \textbf{1.965}& \textbf{44.99}& \textbf{2.751}& \textbf{72.14}& \textbf{2.066}& \textbf{42.77} & $\downarrow$ 25.2\% \\ 
\midrule
Mistral Large & \textbf{0.005}& \ \ \textbf{0.27}& 2.385& 52.45& 2.712& 84.62& 2.209& 47.44 &\\ 
Mistral Large \textsc{(CoT)} & \underline{0.018}& \ \ \underline{0.73}& 2.311& \underline{51.82}& \underline{2.608}& \underline{84.08}& 2.139& \underline{47.12}& $\downarrow$ \ \ 0.7\% \\ 
Mistral Large \textsc{(T\textsuperscript{3}{m})} & 0.039& \ \ 1.84& 2.223& 52.70 & 3.479& 86.18 & 2.399& 48.36& $\uparrow$ \ \ 1.9\% \\ 
Mistral Large \textsc{(T\textsuperscript{3}{d})}  & 0.142&  \ \ 6.59& \underline{1.677}& 57.46 & 2.735& 84.81 & \underline{1.865}& 51.58& $\uparrow$ \ \ 8.7\% \\ 
\rowcolor[gray]{0.9}
Mistral Large \textsc{(T\textsuperscript{3})}& 0.191& \ \ 8.82& \textbf{1.596}& \textbf{42.37}& \textbf{2.137}& \textbf{69.23}& \textbf{1.631}& \textbf{40.70}& $\downarrow$ 14.2\% \\ 
\midrule
GPT-4 & 0.156& \ \ 4.64& {1.167}& 46.05 & 4.114& 88.53 & 2.273& 46.32& \\ 
GPT-4 \textsc{(CoT)} & 0.154& \ \ 4.38& 1.163& {45.86}& 3.981& 88.73 & 2.225& {46.20}& $\downarrow$ \ \ 0.3\% \\ 
GPT-4 \textsc{(T\textsuperscript{3}{m})} & \textbf{0.003}& \ \ \textbf{0.13}& 1.419& 43.22& 2.458& 82.76& 1.621& 42.34&$\downarrow$  \ \ 8.6\% \\ 
GPT-4 \textsc{(T\textsuperscript{3}{d})} & 0.087&\ \ 4.47& \underline{1.124}& \underline{40.13}& \underline{232}& \underline{81.45}& \underline{1.418}& \underline{41.55}&$\downarrow$ 10.3\% \\ 
\rowcolor[gray]{0.9}
GPT-4 \textsc{(T\textsuperscript{3})} & \underline{0.056}& 
 \ \ \underline{3.18}& \textbf{0.854}& \textbf{25.83}& \textbf{1.219}& \textbf{46.22}& \textbf{0.929}& \textbf{25.27}&$\downarrow$ 45.4\% \\ 
\midrule
Claude 3 Opus  & 0.078& \ \ \underline{2.52}& 1.617& 51.36 & 3.713& 88.06 & 2.253& 48.33&\\
Claude 3 Opus \textsc{(CoT)} & \textbf{0.040}& \ \ \textbf{1.59}& 1.642& 49.60 & 3.265 & 87.87 & {2.079}& 47.17 &$\downarrow$ \ \ 2.4\% \\ 
Claude 3 Opus \textsc{(T\textsuperscript{3}m)} & 1.244& 15.14& 1.610& 52.97 & 3.625& 85.27 & 2.426& 51.59&$\uparrow$ \ \ 6.7\% \\ 
Claude 3 Opus \textsc{(T\textsuperscript{3}{d})} & 0.327& \ \ 9.66 & \underline{1.315}& \underline{46.43}& \underline{1.924}& \underline{79.50}& \underline{1.432}& \underline{45.50}&$\downarrow$ \ \ 5.9\% \\ 
\rowcolor[gray]{0.9}
Claude 3 Opus \textsc{(T\textsuperscript{3})}& \underline{0.081}& \ \ 5.30& \textbf{0.406}& \textbf{14.79}& \textbf{0.477}& \textbf{21.29}& \textbf{0.438}& \textbf{14.04}&$\downarrow$ 70.9\% \\ 
            \bottomrule
    \end{tabular}
    \end{center}
        \caption{The comprehensive ablation study results comparing the performance of different prompting methods. We \textbf{bold} the best results and \underline{underline} the second-best results.}
    \label{tab:full_ablation}
\end{table*}
In this section, we extend Section~\ref{sec:ablation_study} and provide a comprehensive supplementary ablation study. We apply two variant methods, \textsc{T\textsuperscript{3}-Merged} \textsc{(T\textsuperscript{3}{m})} and \textsc{T\textsuperscript{3}-Direct-Execution} \textsc{(T\textsuperscript{3}d)}, proposed in Section~\ref{sec:ablation_study}, to all four LLMs capable of applying the \textsc{T}\textsuperscript{3} method. The results of all experiments are listed in Table~\ref{tab:full_ablation}. In this table, we also provide a separate listing for each model, showcasing the average change in error rate for each method compared to the baseline. It is important to note that a lower error rate indicates better performance. 

\paragraph{Impact of \textsc{CoT} Prompting}
Firstly, we observe that \textsc{CoT} prompting yields positive effects across all four LLMs, resulting in an overall reduction in error rate ranging from 0.3\% to 2.4\%. This further corroborates that \textsc{CoT} is a concise and effective prompting strategy.

\paragraph{Impact of \textsc{T\textsuperscript{3}{m}} Prompting}
After applying the \textsc{T\textsuperscript{3}{m}} method, it is observed that the performance of the Mistral Large and Claude 3 Opus deteriorates, particularly with a significant increase in error rate of 6.7\% for Claude 3 Opus. This indicates that the models' ability to summarize the information internally is superior to the approach of extracting and merging information separately. On the contrary, GPT-4 demonstrates a significant reduction in error rate of 8.6\% after applying \textsc{T\textsuperscript{3}{m}}, which indicates that the capabilities of GPT-4 are sufficiently robust to support the \textsc{T\textsuperscript{3}{m}} prompting approach and effectively execute all three steps. The change in error rate for Claude 2.1 is not significant; however, there is a notable 9.6\% reduction in RMSE. This indicates that \textsc{T\textsuperscript{3}{m}} leads to a closer approximation of the true values in its results.

\paragraph{Impact of \textsc{T\textsuperscript{3}{d}} Prompting}

After applying \textsc{T\textsuperscript{3}{d}}, the error rates of Claude 2.1 and Mistral Large increase by 14.1\% and 8.7\%, respectively, indicating that they are not suitable for this prompting method. Comparing these results to those obtained with \textsc{T\textsuperscript{3}}, it can be inferred that the reason for the increase in error rate is likely due to their inferior ability to integrate information compared to the generated code, as the error rate significantly decreases when using code integration. On the other hand, GPT-4 and Claude 3 Opus achieved error rate reductions of 10.3\% and 5.9\%, respectively, under this prompting approach. This indicates that both models possess a certain level of ability to integrate information.

\paragraph{Impact of \textsc{T\textsuperscript{3}} Prompting}

All four LLMs showed substantial improvements after applying the \textsc{T\textsuperscript{3}} prompting method. Specifically, Mistral Large, Claude 2.1, and GPT-4 achieve average error rate reductions of 14.2\%, 25.2\%, and 45.4\% respectively. Notably, Claude 3 Opus exhibits a remarkable error rate reduction of 70.9\%. This strongly indicates the effectiveness of the proposed method. From the perspective of absolute error rates, under the \textsc{T\textsuperscript{3}} method, the performance of the four models from best to worst is as follows: Claude 3 Opus, GPT-4, Mistral Large, and Claude 2.1. This ranking essentially reflects their capabilities in tuple extraction.

\section{Details of the Experiment on \textsc{Struct-Bench} Table Dataset}
\label{sec:app:details_strucbench}

This section is an extension of Section~\ref{sec:struct_bench}, in which we will introduce the evaluation criteria for this dataset, our implementation details, and an analysis of the results.

\begin{table*}[!t]
    \renewcommand\arraystretch{1.06}
\scriptsize
    \centering
    \begin{tabular}{m{1.7cm}m{1.2cm}<{\centering}m{1.2cm}<{\centering}m{1.2cm}<{\centering}m{1.2cm}<{\centering}m{1.2cm}<{\centering}m{1cm}<{\centering}m{1cm}<{\centering}m{1cm}<{\centering}m{1cm}<{\centering}}
    \toprule
       \textbf{Model} &    \textbf{SacreBLEU} &  \textbf{ROUGE-L} & \textbf{BERTScore} & \textbf{BLEURT} & \textbf{BARTScore} & \textbf{Content P-Score} & \textbf{Format P-Score} &  \textbf{Content H-Score} & \textbf{Format H-Score} \\
       \midrule
        \multicolumn{9}{l}{$\vardiamondsuit$ \emph{\textbf{Zero-Shot}}} \\
        ChatGPT & 77.58 & 86.11 & 96.75 & 64.66 & -2.08 & 6.84 & 9.70 & 1.66 & 3.28 \\ 
        \rowcolor[gray]{0.9}
        ChatGPT \textsc{(T\textsuperscript{3})} & 78.91 & 88.36 & 97.34 & 67.47 & -1.90 & 7.29 & \textbf{9.88} & 1.68 & 3.63 \\ 
        GPT-4 &  87.26 & \textbf{92.81} & 98.15 & \underline{77.08} & \underline{-1.58} & 7.45 & 9.71 & \underline{1.76} & \underline{3.87} \\
        \rowcolor[gray]{0.9} 
        GPT-4 \textsc{(T\textsuperscript{3})} & \underline{87.75} & \underline{92.37} & \underline{98.30}  & \textbf{79.80} & -1.60& \underline{7.60} & \underline{9.74} & \textbf{1.77} & \textbf{3.89} \\ 
        \midrule
        \multicolumn{9}{l}{$\vardiamondsuit$ \emph{\textbf{Fine-Tuning}}} \\ 
       LLaMA-7B & \textbf{90.60} &88.98 & \textbf{98.54} & 66.07 & \textbf{-0.69} & \textbf{7.69} & 8.60 & 1.65 & 3.61 \\ 
    \bottomrule
    \end{tabular}
    \caption{Evaluation results on the test set of \textsc{Struc-Bench} Table dataset, consisting of nine metrics. We \textbf{bold} the best results and \underline{underline} the second-best results.}
    \vspace{-0.15in}
    \label{tab:full_struc_bench}
\end{table*}

\subsection{Evaluation Metrics}
This benchmark employs nine evaluation metrics, five of which are traditional and applicable to text generation tasks: {SacreBLEU}~\cite{post-2018-call}, {ROUGE-L}~\cite{lin-2004-rouge}, {BERTScore}~\cite{bert-score}, {BLEURT}~\cite{sellam-etal-2020-bleurt}, and {BARTScore}~\cite{yuan2021bartscore}. These metrics can to some extent measure the similarity between the generated tables and the target tables. The last four metrics, P-score (Prompting Score) and H-score (Heuristical Score), are evaluation criteria proposed by \citet{tang2023strucbench}, which involve using ChatGPT to score the content and format of generated tables and applying manually devised rules to assess the content and format, respectively. For all evaluation metrics, higher numbers signify better performance.

\subsection{Implementation Details}

We conduct experiments on \textsc{Struct-Bench} Table dataset~\cite{tang2023strucbench} using ChatGPT and GPT-4 to assess the impact of \textsc{T}\textsuperscript{3} on model performance. As described in Section~\ref{sec:struct_bench}, we first obtain the outputs of ChatGPT and GPT-4 using the provided prompting, removed terms such as ``unknown'' and ``not mentioned'' from the outputs, and then recalculate all metrics. This step involves using all instructions, prompting templates, and the code for calculating all metrics, all of which are sourced from the codebase\footnote{\href{https://github.com/gersteinlab/Struc-Bench}{https://github.com/gersteinlab/Struc-Bench}} provided by \citet{tang2023strucbench}. We then apply our proposed \textsc{T}\textsuperscript{3} method to both ChatGPT and GPT-4. We design the prompting template for the first step as follows:

\begin{tcolorbox}[title={\textsc{T}\textsuperscript{3} on \textsc{Struct-Bench} Table Dataset \\ Text-to-Tuple Prompting Template}, colback = cBlue_1!10, colframe = cBlue_6,  coltitle=white,fonttitle=\bfseries\small, center title,fontupper=\small]
\texttt{<Instruction>}

You are now required to extract team and player information from the following input. Please focus on the table format and extract all relevant tuples in (\textit{team or player name}, \textit{attribute}, \textit{value}) format:

\texttt{<Text>}
\end{tcolorbox}
\noindent where \texttt{<Instruction>} is provided from the dataset, and \texttt{<text>} is the original text. Because the content of this dataset does not involve information integration, the second step does not alter the tuples from the first step. Therefore, we proceed directly to the third step of table generation, setting the following prompting template:

\begin{tcolorbox}[title={\textsc{T}\textsuperscript{3} on \textsc{Struct-Bench} Table Dataset \\ Tuple-to-Table Prompting Template}, colback = cBlue_1!10, colframe = cBlue_8,  coltitle=white,fonttitle=\bfseries\small, center title,fontupper=\small]

\texttt{<Instruction>}

Based on the instructions and the following extracted tuples, please generate two tables according to the table format:

\texttt{<Tuples>}
\end{tcolorbox}
\noindent where \texttt{<Instruction>} is provided in the dataset, and \texttt{<Tuples>} is the tuples extracted in the previous step.

\subsection{Results Analysis}
Table~\ref{tab:full_struc_bench} presents all our experimental results, with the fine-tuning section quoting the state-of-the-art results~\cite{tang2023strucbench}. The results indicate that GPT-4 outperforms ChatGPT across all metrics. After applying the \textsc{T}\textsuperscript{3} method, both ChatGPT and GPT-4 show improved performance, with ChatGPT experiencing a greater enhancement, suggesting the generalizability of the \textsc{T}\textsuperscript{3} method. When compared with state-of-the-art fine-tuning methods, the zero-shot approaches perform better on some metrics. This indicates that the \textsc{T}\textsuperscript{3} method still offers significant improvements for text-to-table tasks that do not require information integration, further demonstrating its broad applicability.

\section{Details of the Experiment on \textsc{Wiki40b} Dataset}
\label{sec:app:details_wiki40b}
This section is an extension of Section~\ref{sec:struct_sum}, in which we will discuss more details.

\subsection{Evaluation Metrics}
As introduced in Section~\ref{sec:struct_sum}, the evaluation criterion adopted for this dataset, \textsc{Auto-QA} Coverage, is designed due to the absence of ground truth for the generated tables. \citet{jain2024structsum} propose using question answering as a medium and leverage LLMs to assess the quality of the generated tables. This metric not only measures how much information from the original text is covered by the table but also checks the accuracy of the values within the table since incorrect content would also lead to errors in the QA. They also demonstrate through detailed experiments that \textsc{Auto-QA} Coverage aligns with human evaluation standards, and thus we follow this metric in our experiments.

\subsection{Implementation Details}
Since \citet{jain2024structsum} samples 100 entries from the English version of \textsc{Wiki40b}~\cite{49029wiki40B} without disclosing which ones, we replicate their setting by sampling 500 texts to serve as the dataset for this experiment. For each passage, we utilize ChatGPT to generate 20 (question, answer) pairs that can be answered based on the text. Subsequently, we introduced a verification step where all (question, answer) pairs are re-fed into ChatGPT to confirm their correctness. Pairs that ChatGPT cannot accurately answer are filtered out. This approach enhances the accuracy of metric evaluation. Here, we present the prompts for \textsc{T}\textsuperscript{3}:

\begin{tcolorbox}[title={\textsc{T}\textsuperscript{3} on \textsc{Wiki40b} Dataset \\ Text-to-Tuple Prompting Template}, colback = cBlue_1!10, colframe = cBlue_6,  coltitle=white,fonttitle=\bfseries\small, center title,fontupper=\small]
You are going to summarize a table for this passage, but the first step is to extract useful information. Output them in (\textit{subject}, \textit{attribute}, \textit{value}) or (\textit{subject}, \textit{verb}, \textit{object}) format: \texttt{<Text>}
\end{tcolorbox}
\noindent where \texttt{<text>} is the original passage. It is noteworthy that, under this task setting, there are no instructions provided. Next, we feed the output tuples into the second step, where the LLM autonomously integrates the tuples:

\begin{tcolorbox}[title={\textsc{T}\textsuperscript{3} on \textsc{Wiki40b} Dataset \\ Information Integration Prompting Template }, colback = cBlue_1!10, colframe = cBlue_9!70!black,  coltitle=white,fonttitle=\bfseries\small, center title,fontupper=\small]

Please integrate these tuples if necessary: \texttt{<Tuples>}
\end{tcolorbox}
\noindent Finally, the tuples output from this step are fed into the prompt for the last step.

\begin{tcolorbox}[title={\textsc{T}\textsuperscript{3} on \textsc{Wiki40b} Dataset \\ Tuple-to-Table Prompting Template}, colback = cBlue_1!10, colframe = cBlue_8,  coltitle=white,fonttitle=\bfseries\small, center title,fontupper=\small]

Summarize the triples below in one or multiple tables. Use the following format: Caption: A caption for the table you generate. It can be multiple lines. Table: A table in markdown format.

\texttt{<Tuples>}
\end{tcolorbox}
\vspace{-0.05in}

\noindent The other prompts used in the experiments, such as those for generating question-answer pairs, assessing the correctness of answers, and the Divide-and-Generate method's prompting, all originate from \citet{jain2024structsum}.

\subsection{Results Analysis}
The results of this experiment are presented in Figure~\ref{fig:auto_qa_cover}. As discussed in Section~\ref{sec:struct_sum}, \textsc{T}\textsuperscript{3} demonstrates higher \textsc{Auto-QA} Coverage than the baseline, proving its ability to generate higher quality tables in text-to-table generation tasks without instructions. We also experiment with using the tuples extracted in the first step of \textsc{T}\textsuperscript{3} directly as the generated tables and assessed their coverage, resulting in constructive findings. The coverage curve of \textsc{T}\textsuperscript{2} is entirely above that of \textsc{T}\textsuperscript{3}, indicating a certain loss of information from tuple to table. Although this could also be due to insufficient table question answering capabilities leading to a decrease in metrics, it is necessary to employ relevant techniques~\cite{wu-etal-2023-tacr,wang2024chainoftable} to mitigate the impact of this factor in future work. Nevertheless, the conclusion that \textsc{T}\textsuperscript{3} improves the performance of the previous work remains sound.

\subsection{Error Analysis}
We carry out a supplementary case study on the \textsc{Wiki40B} dataset. We sample 50 articles and observe the following situations:
\begin{itemize}
    \item In the first phase, the average accuracy of the tuples obtained from the text-to-tuple process is 81.6\%. We define erroneous tuples as those containing: (i) direct extractions of original sentences, (ii) incomplete triples, and (iii) duplicate triples.
    \item In the second phase, tuple integration, 24\% of the articles undergo integration of similar items, while the remainder simply involves sorting and categorizing the tuples or outputting them unchanged.
    \item In the third phase, tuple-to-table essentially involves treating each tuple as a row to be filled into a table, but two scenarios may introduce errors: (i) 14\% of the articles involve the fragmentation of long sentences, leaving only incomplete tuples, and (ii) 6\% of the articles continue to merge similar items based on the tuples, combining related tuples into long sentences. (While this step may not necessarily introduce errors, it could increase the difficulty of reasoning, and the \textsc{AutoQA} metric may decrease.)
\end{itemize}

\section{Case Studies}
\label{sec:app:case_study}

Figure~\ref{fig:case_study} lists the outputs of four LLMs with and without the application of the \textsc{T}\textsuperscript{3} method on the data shown in Figure~\ref{fig:full_example}. For the results not utilizing \textsc{T}\textsuperscript{3}, we can not easily analyze why it generates a range of large or small values. However, for the results using the \textsc{T}\textsuperscript{3} method, we perform a detailed examination. We randomly sample 100 results generated by GPT-4 applying \textsc{T}\textsuperscript{3} and conduct a spot check, finding that all errors originated from the first stage. Among these errors, 78\% are due to missing event tuples, and 21\% are due to wrong event tuples. Here, we present two representative examples.

\vspace{-0.05in}
\begin{tcolorbox}[title={Missing Event Tuple Example}, colback = cBlue_1!10, colframe = cBlue_9,  coltitle=white,fonttitle=\bfseries\small, center title,fontupper=\small,fontlower=\small]

Player18(Home Team) earns a free kick on the left wing after being fouled by Player20(Away Team). 

\tcblower

\texttt{(Player18, Home Team, Free Kick)}

\textcolor{gray}{\texttt{(Player20, Away Team, Foul)}(missing)}

\end{tcolorbox}
\vspace{-0.05in}
\begin{tcolorbox}[title={Wrong Event Tuple Example}, colback = cBlue_1!10, colframe = cBlue_9,  coltitle=white,fonttitle=\bfseries\small, center title,fontupper=\small,fontlower=\small]

Player17(Home Team) from the Home Team draws a foul in the penalty area, resulting in a penalty conceded by Player33(Away Team).

\tcblower

\textcolor{cRed}{\texttt{(Player17, Home Team, Foul)}(wrong)}

\texttt{(Player33, Away Team, Foul)}

\textcolor{gray}{\texttt{(Player17, Home Team, Free Kick)}(missing)}

\end{tcolorbox}
\vspace{-0.05in}

\begin{figure*}[!t]

\vspace{-0.05in}
\begin{tcolorbox}[title={Outputs from Claude 2.1}, colback = cBlue_1!10, colframe = cBlue_6!80,  coltitle=white,fonttitle=\bfseries\small,fontupper=\scriptsize,fontlower=\scriptsize]

\textbf{Results w/o \textsc{T}\textsuperscript{3}:} RMSE=1.888, Error Rate=43.75\%.

\centering
\begin{tabular}{m{1.2cm}<{\centering}m{1cm}<{\centering}m{1cm}<{\centering}m{1cm}<{\centering}m{1.5cm}<{\centering}m{1.2cm}<{\centering}m{1.5cm}<{\centering}m{1.2cm}<{\centering}m{1cm}<{\centering}}
    \toprule
        \textbf{Team} &  \textbf{Goals} & \textbf{Shots} & \textbf{Fouls} & \textbf{Yellow Cards} & \textbf{Red Cards} & \textbf{Corner Kicks} & \textbf{Free Kicks} & \textbf{Offsides} \\
        \midrule
         Away Team & 3 \ding{51} & 13 \ding{55}  & 10 \ding{55} & 0 \ding{51} & 0 \ding{51} & 3 \ding{51} & 5 \ding{55} & 4 \ding{55} \\
         \midrule
         Home Team & 0 \ding{51}  & 10 \ding{55}  & 7 \ding{55} & 1 \ding{51} & 0 \ding{51} & 5 \ding{51} & 6 \ding{51} & 3 \ding{55}\\
         \bottomrule
    \end{tabular}
\tcblower
\textbf{Results w/ \textsc{T}\textsuperscript{3}:} RMSE=0.661, Error Rate=25.00\%.

\centering
\begin{tabular}{m{1.2cm}<{\centering}m{1cm}<{\centering}m{1cm}<{\centering}m{1cm}<{\centering}m{1.5cm}<{\centering}m{1.2cm}<{\centering}m{1.5cm}<{\centering}m{1.2cm}<{\centering}m{1cm}<{\centering}}
    \toprule
        \textbf{Team} &  \textbf{Goals} & \textbf{Shots} & \textbf{Fouls} & \textbf{Yellow Cards} & \textbf{Red Cards} & \textbf{Corner Kicks} & \textbf{Free Kicks} & \textbf{Offsides} \\
                 \midrule
         Away Team & 3 \ding{51} & 12 \ding{51}  & 4 \ding{55} & 0 \ding{51} & 0 \ding{51} & 3 \ding{51} & 6 \ding{51} & 2 \ding{51} \\
         \midrule
         Home Team & 0 \ding{51}  & 5 \ding{51}  & 7 \ding{55} & 1 \ding{51} & 0 \ding{51} & 6 \ding{55} & 6 \ding{51} & 5 \ding{55}\\
         \bottomrule
    \end{tabular}
\end{tcolorbox}
\vspace{-0.05in}

\vspace{-0.05in}
\begin{tcolorbox}[title={Outputs from Mistral Large}, colback = cBlue_1!10, colframe = cBlue_6!70,  coltitle=white,fonttitle=\bfseries\small,fontupper=\scriptsize,fontlower=\scriptsize]

\textbf{Results w/o \textsc{T}\textsuperscript{3}:} RMSE=1.458, Error Rate=50.00\%.

\centering
\begin{tabular}{m{1.2cm}<{\centering}m{1cm}<{\centering}m{1cm}<{\centering}m{1cm}<{\centering}m{1.5cm}<{\centering}m{1.2cm}<{\centering}m{1.5cm}<{\centering}m{1.2cm}<{\centering}m{1cm}<{\centering}}
    \toprule
        \textbf{Team} &  \textbf{Goals} & \textbf{Shots} & \textbf{Fouls} & \textbf{Yellow Cards} & \textbf{Red Cards} & \textbf{Corner Kicks} & \textbf{Free Kicks} & \textbf{Offsides} \\
         \midrule
         Away Team & 3 \ding{51} & 10 \ding{55}  & 7 \ding{55} & 0 \ding{51} & 0 \ding{51} & 3 \ding{51} & 10 \ding{55} & 2 \ding{51} \\
         \midrule
         Home Team & 0 \ding{51}  & 8 \ding{55}  & 7 \ding{55} & 1 \ding{51} & 0 \ding{51} & 4 \ding{55} & 7 \ding{55} & 5 \ding{55}\\
         \bottomrule
    \end{tabular}
\tcblower
\textbf{Results w/ \textsc{T}\textsuperscript{3}:} RMSE=0.612, Error Rate=18.75\%.

\centering
\begin{tabular}{m{1.2cm}<{\centering}m{1cm}<{\centering}m{1cm}<{\centering}m{1cm}<{\centering}m{1.5cm}<{\centering}m{1.2cm}<{\centering}m{1.5cm}<{\centering}m{1.2cm}<{\centering}m{1cm}<{\centering}}
    \toprule
        \textbf{Team} &  \textbf{Goals} & \textbf{Shots} & \textbf{Fouls} & \textbf{Yellow Cards} & \textbf{Red Cards} & \textbf{Corner Kicks} & \textbf{Free Kicks} & \textbf{Offsides} \\
                 \midrule
         Away Team & 3 \ding{51} & 12 \ding{51}  & 6 \ding{51} & 0 \ding{51} & 0 \ding{51} & 3 \ding{51} & 4 \ding{55} & 2 \ding{51} \\
         \midrule
         Home Team & 0 \ding{51}  & 5 \ding{51}  & 5 \ding{55} & 1 \ding{51} & 0 \ding{51} & 5 \ding{51} & 5 \ding{55} & 6 \ding{51}\\
         \bottomrule
    \end{tabular}
\end{tcolorbox}
\vspace{-0.05in}

\vspace{-0.05in}
\begin{tcolorbox}[title={Outputs from GPT-4}, colback = cBlue_1!10, colframe = cBlue_6!60,  coltitle=white,fonttitle=\bfseries\small,fontupper=\scriptsize,fontlower=\scriptsize]

\textbf{Results w/o \textsc{T}\textsuperscript{3}:} RMSE=1.785, Error Rate=31.25\%.

\centering
\begin{tabular}{m{1.2cm}<{\centering}m{1cm}<{\centering}m{1cm}<{\centering}m{1cm}<{\centering}m{1.5cm}<{\centering}m{1.2cm}<{\centering}m{1.5cm}<{\centering}m{1.2cm}<{\centering}m{1cm}<{\centering}}
    \toprule
        \textbf{Team} &  \textbf{Goals} & \textbf{Shots} & \textbf{Fouls} & \textbf{Yellow Cards} & \textbf{Red Cards} & \textbf{Corner Kicks} & \textbf{Free Kicks} & \textbf{Offsides} \\
        \midrule
         Away Team & 3 \ding{51} & 10 \ding{55}  & 6 \ding{51} & 0 \ding{51} & 0 \ding{51} & 3 \ding{51} & 5 \ding{55} & 2 \ding{51} \\
         \midrule
         Home Team & 0 \ding{51}  & 11 \ding{55}  & 9 \ding{55} & 1 \ding{51} & 0 \ding{51} & 5 \ding{51} & 7 \ding{55} & 6 \ding{51}\\
         \bottomrule
    \end{tabular}
\tcblower
\textbf{Results w/ \textsc{T}\textsuperscript{3}:} RMSE=0.433, Error Rate=18.75\%.

\centering
\begin{tabular}{m{1.2cm}<{\centering}m{1cm}<{\centering}m{1cm}<{\centering}m{1cm}<{\centering}m{1.5cm}<{\centering}m{1.2cm}<{\centering}m{1.5cm}<{\centering}m{1.2cm}<{\centering}m{1cm}<{\centering}}
    \toprule
        \textbf{Team} &  \textbf{Goals} & \textbf{Shots} & \textbf{Fouls} & \textbf{Yellow Cards} & \textbf{Red Cards} & \textbf{Corner Kicks} & \textbf{Free Kicks} & \textbf{Offsides} \\
        \midrule
         Away Team & 3 \ding{51} & 12 \ding{51}  & 5 \ding{55} & 0 \ding{51} & 0 \ding{51} & 3 \ding{51} & 5 \ding{55} & 2 \ding{51} \\
         \midrule
         Home Team & 0 \ding{51}  & 5 \ding{51}  & 6 \ding{51} & 1 \ding{51} & 0 \ding{51} & 5 \ding{51} & 5 \ding{55} & 6 \ding{51}\\
         \bottomrule
    \end{tabular}
\end{tcolorbox}
\vspace{-0.05in}

\vspace{-0.05in}
\begin{tcolorbox}[title={Outputs from Claude 3 Opus}, colback = cBlue_1!10, colframe = cBlue_6!50,  coltitle=white,fonttitle=\bfseries\small,fontupper=\scriptsize,fontlower=\scriptsize]

\textbf{Results w/o \textsc{T}\textsuperscript{3}:} RMSE=2.046, Error Rate=56.25\%.

\centering
\begin{tabular}{m{1.2cm}<{\centering}m{1cm}<{\centering}m{1cm}<{\centering}m{1cm}<{\centering}m{1.5cm}<{\centering}m{1.2cm}<{\centering}m{1.5cm}<{\centering}m{1.2cm}<{\centering}m{1cm}<{\centering}}
    \toprule
        \textbf{Team} &  \textbf{Goals} & \textbf{Shots} & \textbf{Fouls} & \textbf{Yellow Cards} & \textbf{Red Cards} & \textbf{Corner Kicks} & \textbf{Free Kicks} & \textbf{Offsides} \\
        \midrule
         Away Team & 3 \ding{51} & 18 \ding{55}  & 7 \ding{55} & 0 \ding{51} & 0 \ding{51} & 2 \ding{55} & 7 \ding{55} & 3 \ding{55} \\
         \midrule
         Home Team & 0 \ding{51}  & 9 \ding{55}  & 7 \ding{55} & 1 \ding{51} & 0 \ding{51} & 5 \ding{51} & 9 \ding{55} & 5 \ding{55}\\
         \bottomrule
    \end{tabular}
\tcblower
\textbf{Results w/ \textsc{T}\textsuperscript{3}:} RMSE=0.000, Error Rate=0.00\%.

\centering
\begin{tabular}{m{1.2cm}<{\centering}m{1cm}<{\centering}m{1cm}<{\centering}m{1cm}<{\centering}m{1.5cm}<{\centering}m{1.2cm}<{\centering}m{1.5cm}<{\centering}m{1.2cm}<{\centering}m{1cm}<{\centering}}
    \toprule
        \textbf{Team} &  \textbf{Goals} & \textbf{Shots} & \textbf{Fouls} & \textbf{Yellow Cards} & \textbf{Red Cards} & \textbf{Corner Kicks} & \textbf{Free Kicks} & \textbf{Offsides} \\
        \midrule
         Away Team & 3 \ding{51} & 12 \ding{51}  & 6 \ding{51} & 0 \ding{51} & 0 \ding{51} & 3 \ding{51} & 6 \ding{51} & 2 \ding{51} \\
         \midrule
         Home Team & 0 \ding{51}  & 5 \ding{51}  & 6 \ding{51} & 1 \ding{51} & 0 \ding{51} & 5 \ding{51} & 6 \ding{51} & 6 \ding{51}\\
         \bottomrule
    \end{tabular}
\end{tcolorbox}
\vspace{-0.05in}
\caption{Case study analysis showing the outputs and evaluation metrics of Claude 2.1, Mistral Large, GPT-4, and Claude 3 Opus with and without the \textsc{T}\textsuperscript{3} method on data shown in Figure~\ref{fig:full_example}.}
\label{fig:case_study}
\end{figure*}

\end{document}